\documentclass[lettersize,journal]{IEEEtran}
\usepackage{amsmath,amsfonts}
\usepackage{algorithmic}
\usepackage{algorithm}
\usepackage{array}
\usepackage[caption=false,font=normalsize,labelfont=sf,textfont=sf]{subfig}
\usepackage{textcomp}
\usepackage{stfloats}
\usepackage{url}
\usepackage{verbatim}
\usepackage{bm}
\usepackage{graphicx}
\usepackage{bbm}
\usepackage{cite}
\usepackage{pifont}
\usepackage{amssymb}
\usepackage{adjustbox}
\usepackage{booktabs}
\usepackage{multirow}
\usepackage{makecell}
\usepackage[table]{xcolor}
\definecolor{softgray}{gray}{0.85}
\definecolor{softblue}{RGB}{220, 235, 250}  
\definecolor{softgreen}{RGB}{220, 245, 220} 
\begin{document}

\title{PDAC: Efficient Coreset Selection for Continual Learning via Probability Density Awareness}

\author{Junqi Gao\textsuperscript{1}, Zhichang Guo\textsuperscript{1}, Dazhi Zhang\textsuperscript{1}, Yao Li\textsuperscript{1}, Yi Ran\textsuperscript{1}, Biqing Qi\textsuperscript{2}
        % <-this % stops a space
        
\thanks{Corresponding authors: Zhichang Guo and Dazhi Zhang.

\textsuperscript{1}School of Mathematics, Harbin Institute of Technology, Harbin, P. R. China;
\textsuperscript{2}Shanghai Artificial Intelligence Laboratory, Shanghai, P. R. China.
(Emails: gjunqi97@gmail.com; mathgzc@hit.edu.cn; zhangdazhi@hit.edu.cn; yaoli0508@hit.edu.cn; 21b912025@stu.hit.edu.cn; qibiqing7@gmail.com).

}

}

\maketitle

\begin{abstract}
Rehearsal-based Continual Learning (CL) maintains a limited memory buffer to store replay samples for knowledge retention, making these approaches heavily reliant on the quality of the stored samples. Current Rehearsal-based CL methods typically construct the memory buffer by selecting a representative subset (referred to as coresets), aiming to approximate the training efficacy of the full dataset with minimal storage overhead. However, mainstream Coreset Selection (CS) methods generally formulate the CS problem as a bi-level optimization problem that relies on numerous inner and outer iterations to solve, leading to substantial computational cost thus limiting their practical efficiency. In this paper, we aim to provide a more efficient selection logic and scheme for coreset construction. To this end, we first analyze the Mean Squared Error (MSE) between the buffer-trained model and the Bayes-optimal model through the perspective of localized error decomposition to investigate the contribution of samples from different regions to MSE suppression. Further theoretical and experimental analyses demonstrate that samples with high probability density play a dominant role in error suppression. Inspired by this, we propose the Probability Density-Aware Coreset (PDAC) method. PDAC leverages the Projected Gaussian Mixture (PGM) model to estimate each sample's joint density, enabling efficient density-prioritized buffer selection. Finally, we introduce the streaming Expectation Maximization (EM) algorithm to enhance the adaptability of PGM parameters to streaming data, yielding Streaming PDAC (SPDAC) for streaming scenarios. Extensive comparative experiments show that our methods outperforms other baselines across various CL settings while ensuring favorable efficiency.
\end{abstract}

\begin{IEEEkeywords}
Continual Learning, Coreset Selection, Rehearsal-based methods, Density Estimation
\end{IEEEkeywords}

\section{Introduction}
Deep learning models have demonstrated remarkable performance in various downstream tasks, such as image classification \cite{HeZRS16, DosovitskiyB0WZ21} and object detection \cite{HeGDG17, Girshick15}. However, once these models are trained on new tasks, they often experience rapid loss of knowledge about old tasks, a phenomenon commonly referred to as catastrophic forgetting \cite{french1999catastrophic, EWC}.

To alleviate this issue, Continual Learning (CL) has been introduced to enable models to learn a sequence of tasks while retaining knowledge from previously learned tasks. Among existing approaches, the most straightforward and effective is the rehearsal-based CL method. It maintains a limited memory buffer and performs sample replay based on the memory buffer during the training of subsequent tasks \cite{chaudhry2019tiny,BuzzegaBPAC20}, thereby preserving the model’s memory of historical tasks. 
\begin{figure}[htbp]
    \centering
    \includegraphics[width=0.48\textwidth]{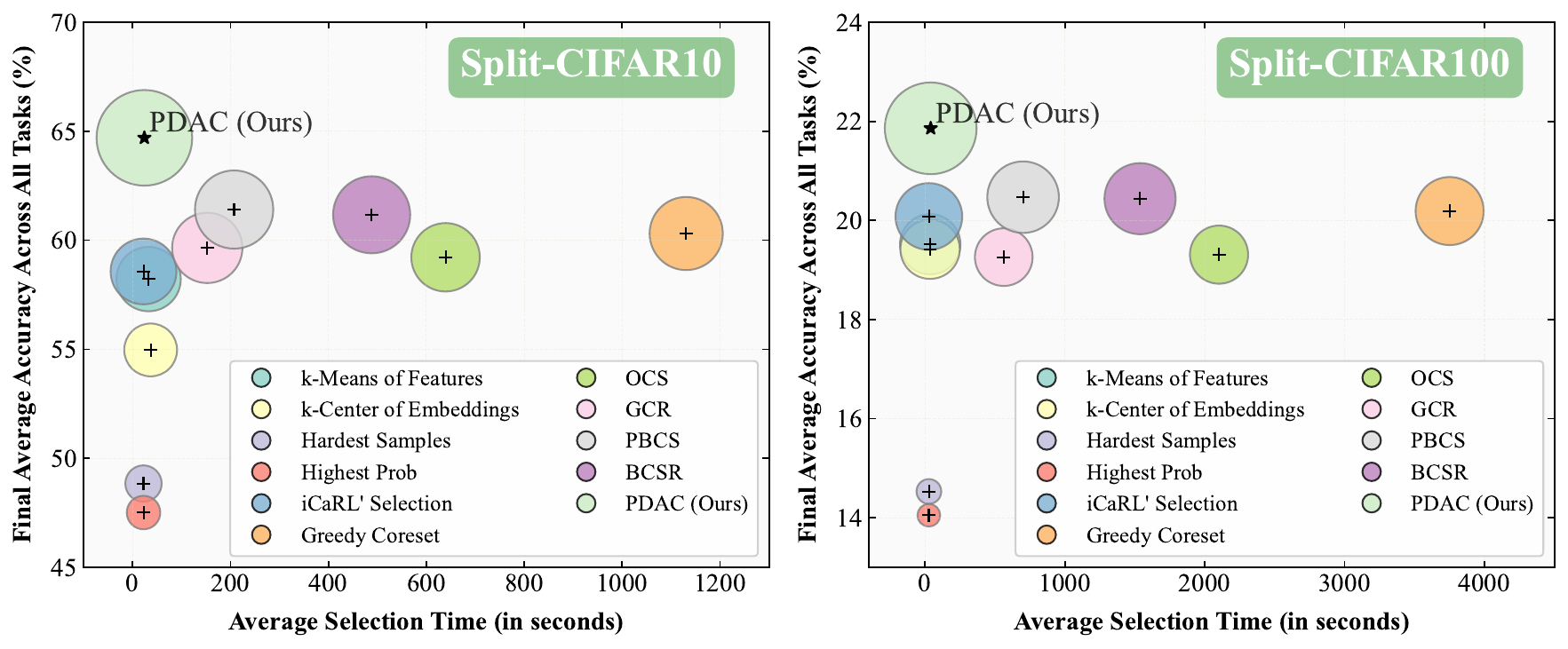}  
    \caption{Comparison of the final average accuracy ($\%$) and average selection time (in seconds) after CL training on Split-CIFAR10 (left) and Split-CIFAR100 (right) for the proposed PDAC against other representative CS methods. The y-coordinate of each bubble's center represents the method's final average accuracy, while its size is positively correlated with the its accuracy. The central marker indicates the method's performance point. PDAC achieves high performance while maintaining good efficiency.}  
    \vspace{-3pt}
    \label{fig:intro}
    \vspace{-5pt}
\end{figure}

For rehearsal-based CL methods, the construction strategy of the memory buffer plays a decisive role in model performance \cite{TiwariKIS22,ShimMJSKJ21}. To build a high-quality buffer, Coreset Selection (CS) is widely adopted, which aims to choose the most informative and representative subset (referred to as the coreset) from the training set of each task, ensuring that training solely on this subset approximates the performance achieved on the full dataset \cite{BorsosM020,YoonMYH22}. Current advanced CS methods typically formulate sample selection as a bi-level optimization problem \cite{ZhouPZLCZ22,HaoJL23}. For each candidate dataset, these methods select samples by alternately performing inner-layer loss optimization and outer-layer sample importance optimization. However, such strategies require up to hundreds of inner and outer iterations during each selection process, along with repeated leave-one-out retraining on the candidate set, which significantly limits the selection efficiency.

To address this efficiency bottleneck, we aim to provide a more direct and efficient selection criterion for constructing high-quality coresets, avoiding the cumbersome leave-one-out validation process. To this end, we first perform a localized decomposition of the Mean Squared Error (MSE) between the model trained on selected buffer samples and the Bayes-optimal model with respect to the sample space. Building on this decomposition, we conduct theoretical analyses of how samples from different local regions contribute to error suppression. Combined with our designed validation experiments, we draw a core conclusion: \textbf{high-quality coreset construction should prioritize samples with high probability densities}. 

Guided by this principle, we introduce the Projected Gaussian Mixture (PGM) model to estimate the real sample density using the output features corresponding to each sample. Specifically, we first apply a Variance Maximizing Projection (VMP) to the output features, reducing their dimensionality while effectively preserving the dominant modes of feature variation, thereby alleviating the challenges posed by high-dimensional feature sparsity for density estimation. Subsequently, we use Gaussian Mixture Models (GMMs) to perform density estimation on the projected features, constructing class-conditional density estimates. By further incorporating empirical class frequencies to estimate class priors, we construct a full joint density estimate for each sample. 

Building on this density estimator, we propose the Probability Density-Aware Coreset Selection (PDAC) method to achieve efficient offline buffer selection. Finally, we introduce the streaming Expectation Maximization (EM) algorithm to enhance the adaptability of PGM parameters to streaming data, and develop the Streaming PDAC (SPDAC) method for memory buffer selection in streaming data scenarios. Comparative experiments under offline setting and the more challenging streaming setting demonstrate that our methods can achieve superior performance compared to baselines while ensuring favorable efficiency.

In summary, our contributions can be summarized as follows:
\begin{itemize}
\item Using localized MSE decomposition, we theoretically analyze how samples from different regions contribute to error reduction and derive a selection criterion that prioritizes samples with higher probability density.
\item We propose PDAC to perform efficient offline buffer selection using PGM density estimates. By incorporating the streaming EM algorithm, we extend it to SPDAC for buffer selection in streaming scenarios.
\item Comparative experiments show that our method outperforms other baselines across various CL settings while maintaining favorable efficiency.
\end{itemize}
\section{Related Works}
\subsection{Coreset Selection}
Coreset Selection (CS) algorithms aim to approximate the training performance on the original dataset based on a certain criterion (e.g., training loss, model gradients, etc.) by selecting a representative data subset. Early CS research primarily focused on various classical supervised learning algorithms, such as support vector machine \cite{tsang2005core} and logistic regression \cite{HugginsCB16,MunteanuSSW18}, as well as unsupervised algorithms like K-means \cite{Har-PeledM04,har2005smaller,FeldmanL11} and GMMs \cite{FeldmanFK11,lucic2018training}. Nguyen et al. \cite{NguyenLBT18} introduced a K-Center-based CS strategy \cite{gonzalez1985clustering} to CL problems, while Yoon et al. \cite{YoonMYH22} and Tiwari et al. \cite{TiwariKIS22} performed CS through model gradient matching. Borsos et al. \cite{BorsosM020} were the first to formulate the CS problem as a bilevel optimization problem, and subsequent work has focused on optimizing the modeling of the bilevel problem, such as introducing a probabilistic framework \cite{ZhouPZLCZ22} or incorporating loss regularization terms \cite{HaoJL23}. Although showing promising performance, solving bi-level optimization incurs excessive computational overhead and over-relies on the model’s current training state, making it vulnerable to fluctuations in training dynamics. In contrast, our approach directly guides sample-level evaluation and selection using adaptively updated sample statistics, eliminating the need for greedy optimization over candidate sets. Thus, it achieves both effectiveness and efficiency.

\subsection{Rehearsal-based Continual Learning}
Rehearsal-based CL methods maintain a limited memory buffer by selecting and storing samples from previous tasks and replaying them during subsequent task training to preserve the model's knowledge of the prior tasks. Classic methods are represented by random selection-based strategies \cite{chaudhry2019tiny,hayes2019memory}, while subsequent approaches further enhance the effectiveness of replay training by incorporating distillation loss \cite{RebuffiKSL17,li2023variational}, classifier output matching \cite{BuzzegaBPAC20,QiGCLLWZ25}, gradient episodic memory \cite{Lopez-PazR17,ChaudhryRRE19}. Among these, CS is widely used as a representative strategy to construct high-quality memory buffers.

\begin{table}[h]
\setlength{\tabcolsep}{3pt}
\centering
\caption{Symbol Specifications}
\label{symbols}
\begin{tabular}{lc} 
\toprule
Symbol & Description \\
\midrule
$\mathcal{X}$ & Input space \\
$\mathcal{Y}$ & Set of labels \\
$ \mathcal{Z}$ & Sample Space \\
$\bm{z}=\left(\bm{x},y\right)$ & A sample with $\bm{x}$ as input and $y$ as label \\
$\mathcal{T}^{t}$ & The $t$-th CL task \\
$\mathcal{S}^t$ & The training set corresponding to task $\mathcal{T}^{t}$\\
$\mathcal{Y}^{t}$ & The set of class labels included in task $\mathcal{T}^{t}$\\
$\mathcal{E}^{t}$ & The test set corresponding to task $\mathcal{T}^{t}$\\
$\bm{z}_{t,i}$ & The $i$-th sample of the $t$-th task's training set $\mathcal{S}^t$\\
$\mathcal{M}$ & Memory buffer \\
$N$ & The size of the memory buffer \\
$n_y$ & Number of training samples of class $y$\\
$p\left(\bm{z}\right)$ & Real probability density of sample $\bm{z}$\\
$\pi\left(z\right)$ & Resampling probability mass of sample $\bm{z}$\\
$f\left(\cdot\,;\bm{\theta}\right)$ & Parameterized learning model with parameter $\bm{\theta}$\\
$\Theta$ & The parameter space \\
$\sigma\left(\cdot\right)$ & Softmax activation function \\
$f^\sigma=\sigma\circ f$ & Model output after Softmax activation \\
${\bm{\hat\theta}}_{\mathcal{M}}$ & The ERM model trained on the memory buffer $\mathcal{M}$\\
$f^*\left(\cdot\right)$ & Bayes-optimal model\\
$\mathcal{R}_{\mathcal{M}\mid\mathcal{S}}$ & MSE between the buffer-trained model and $f^*\left(\cdot\right)$\\
$\mathcal{X}_i$ & The $i$-th local unit of the sample space $\mathcal{X}$\\
$\mathcal{Z}_i$ & The $i$-th local region of the sample space $\mathcal{Z}$\\
$m$ & The diameter of each local unit\\
$\mathrm{p}_i$ & Probability of a resampled point belonging to $\mathcal{Z}_i$\\
$l_i$ & Number of training samples in local region $\mathcal{Z}_i$\\
$\bm{W}_{y}^t$ & VMP projection matrix for class $y$ in task $\mathcal{T}^t$\\
$f_\psi\left(\cdot\,;\bm{\theta}\right)$ & Penultimate layer output of the model $f\left(\cdot\,;\bm{\theta}\right)$\\
$\bm{h}=f_\psi\left(\bm{x};\bm{\theta}\right)$ & Feature of sample $\bm{x}$ extracted by $f_\psi\left(\cdot\,;\bm{\theta}\right)$\\
$\bm{\xi}$ & The projected sample feature\\
$L$ & Number of Gaussian components in PGM\\
$\alpha_y^l$ & Weight of the $l$-th Gaussian component for class $y$\\
$\bm{\mu}_y^l$ & Mean vector of the $l$-th component within class $y$\\
$\bm{\Sigma}_{y}^l$ & Covariance matrix of the $l$-th component within class $y$\\
$\beta$ & Update step size in the streaming EM algorithm\\
$\mathcal{C}_y^t$ & Index set of samples belonging to class $y$ in task $\mathcal{T}^t$\\
$\mathcal{B}$ & Input batch\\
\bottomrule
\end{tabular}
\label{tab:symbol}
\end{table}
\section{Problem Formulation}
The CL problem we consider consists of $T$ sequentially arranged classification tasks. Each task, denoted as $\mathcal T^t$, corresponds to a training set
$\mathcal S^t=\{(\bm x_{t,i}, y_{t,i})\}_{i=1}^{\left|\mathcal S^t\right|}$ and an test set $\mathcal E^t=\{(\bm x^{\mathcal E}_{t,i}, y^{\mathcal E}_{t,i})\}_{i=1}^{\left|\mathcal E^t\right|}$, where $\left|\mathcal S\right|$ represents the number of elements in set $\mathcal S$. 
The set of class labels included in $\mathcal T^t$ is denoted as $\mathcal Y^t$, which satisfies $\mathcal Y^{t_1}\cap \mathcal Y^{t_2} = \emptyset, \forall t_1,t_2\in \{1, 2, \dots, T\}$ and the set of all class labels $\mathcal Y=\cup_{t=1}^{T}\mathcal Y^t$. The model is sequentially trained on tasks $\mathcal T^1, \mathcal T^2, \dots, \mathcal T^T$. During the training of task $\mathcal T^t$, the previous training sets $\mathcal S^1, \mathcal S^2, \dots, \mathcal S^{t-1}$ become inaccessible. Upon completion of the training on task $\mathcal T^t$, the model is evaluated on the union of evaluation sets from all seen tasks $\cup_{i=1}^{t}\mathcal E^i$. The rehearsal-based CL methods maintains a memory buffer $\mathcal M$ of total size $N$ to assist the model in recalling previous knowledge during subsequent tasks. When training on $\mathcal T^t$, one can select samples from the training set $\mathcal S^t$ and add to $\mathcal M$. During each iteration of $\mathcal{T}^{t+1}$, the replayed data can be retrieved from $\mathcal{M}$ and added to the current training batch. Our symbol setting can be summarized in Tab. \ref{tab:symbol}.

\section{The Role of Data Density in Sample Selection}
After establishing the aforementioned problem framework, we turn to a central question of the CS problem: \textit{given a training set, how can we design a sampling strategy that ensures buffer samples more effectively preserve the model’s performance on that data distribution?} Next, we will first conduct theoretical analysis and validation experiments, and then develop a CS strategy for constructing memory buffer based on the derived conclusions. All theoretical proofs are included in the Appendix of the supplementary material.
\subsection{Local Error Decomposition in Buffer Selection}
Let the sample space be $\mathcal{Z} = \mathcal{X} \times \mathcal{Y}$ with the corresponding sample distribution $p(\boldsymbol{z})$, where $\boldsymbol z = (\boldsymbol x, y)$. Denote $\boldsymbol{z}_i=\left(\boldsymbol{x}_i, y_i\right)$, consider an i.i.d. sample set $\mathcal{S} = \{\boldsymbol{z}_i\}_{i=1}^n$ containing $n$ samples from $K$ classes drawn from $p(\boldsymbol{z})$. Secondary sampling can be performed from $\mathcal{S}$ according to the probability mass $\pi\left(\boldsymbol{z}\right)$ to obtain a Memory Buffer $\mathcal{M} = \{\boldsymbol{z}'_i\}_{i=1}^N \subset \mathcal{S}$ containing $N$ ($N \leq n$) samples. By adjusting the values of $\pi\left(\boldsymbol{z}\right)$ in different regions of the input space $\mathcal{X}$, one can achieve localized adjustments to the resampling probabilities, thereby reshaping the data distribution within $\mathcal{M}$.

Given the parameter space $\Theta$ of the parameterized learning model $f(\cdot; \boldsymbol{\theta}): \mathcal{X} \to \mathbb{R}^K$, we denote the Softmax operation as $\sigma(\cdot): \mathbb{R}^K \to [0,1]^K$. For $\boldsymbol{x} \in \mathcal{X}$, the $i$-th component of the Softmax output is defined as
$\sigma(f(\boldsymbol{x}; \boldsymbol{\theta}))_i = \frac{\exp(f(\boldsymbol{x}; \boldsymbol{\theta})_i)}{\sum_{k=1}^K \exp(f(\boldsymbol{x}; \boldsymbol{\theta})_k)}$, here $1 \leq i \leq K$ and $\boldsymbol{v}_k$ represents the $k$-th component of the vector $\boldsymbol{v}$. For simplicity, we denote $f^\sigma = \sigma \circ f$. Define the learning loss as $\ell(\cdot, \cdot): [0,1] \times \mathcal{Y} \to \mathbb{R}^+$. The Empirical Risk Minimization (ERM) model on $\mathcal{M}$ is denoted by $f(\cdot; \boldsymbol{\hat{\theta}}_\mathcal{M}): \mathcal{X} \rightarrow \mathbb{R}^K$, which satisfies $\boldsymbol{\hat{\theta}}_\mathcal{M} = \arg\min_{\boldsymbol{\theta} \in \Theta} \frac{1}{N} \sum_{i=1}^N \ell(f^\sigma(\boldsymbol{x}'_i; \boldsymbol{\theta})_{y'_i}, y'_i)$, here $\left(\boldsymbol{x}'_i,y'_i\right)=\boldsymbol{z}'_i\in\mathcal{M}$. Let $f^*\left(\cdot\right):\mathcal X\to[0,1]^K$ denote the Bayes-optimal model, i.e., the optimal predictor in expectation under no constraints, defined by $f^*=\arg\min_{f\in\mathcal{F}} \mathbb E_{\boldsymbol z}\!\bigl[\ell\bigl(f(\boldsymbol x)_y,y\bigr)\bigr]$, where $\mathcal{F}$ represents the hypothesis class consisting of all $f$ that map from $\mathcal{X}$ to $[0,1]^K$. Considering that $\mathcal{S}$ is typically given, we examine the following Mean Squared Error (MSE) under the condition of a given $\mathcal{S}$:
\begin{equation}
\mathcal{R}_{\mathcal{M} \mid \mathcal{S}} = \mathbb{E}_{\boldsymbol{z}} \mathbb{E}_{\mathcal{M} \mid \mathcal{S}} \left[ \left\| f^\sigma\left(\boldsymbol{x}; \boldsymbol{\hat{\theta}}_\mathcal{M}\right) - f^*\left(\boldsymbol{x}\right) \right\|_2^2 \right],
\end{equation}
where $\mathbb{E}_{\mathcal{M} \mid \mathcal{S}}$ denotes the conditional expectation with respect to $\mathcal{M}$ under the condition that $\mathcal{S}$ is given. The error $\mathcal{R}_{\mathcal{M} \mid \mathcal{S}}$ measures the difference in expectation between the model $ f^\sigma(\cdot; \boldsymbol{\hat{\theta}}_\mathcal{M}) $ trained on $\mathcal{M}$ and the Bayes-optimal model $ f^*(\cdot) $, given $\mathcal{S}$. A smaller $\mathcal{R}_{\mathcal{M} \mid \mathcal{S}}$ indicates that the choice of $\mathcal{M}$ is more conducive to maintaining model behavior that is closer to that of the golden predictor. 

We aim to investigate the local properties of $\mathcal{R}_{\mathcal{M} \mid \mathcal{S}}$ to provide insights into the selection of $\pi(\boldsymbol{z})$ in different regions. To this end, we consider the case of bounded inputs, that is, $\forall \boldsymbol{x} \in \mathcal{X}$, $\|\boldsymbol{x}\|_{\infty} \leq M$ and each component of $\boldsymbol{x}$ is non- negative, which is consistent with the common practice of input normalization. Further, we divide the entire hypercube covering the input space into local units with a diameter of $m$, resulting in a total of $L_{m} = \left(\frac{M}{m}\right)^{d}$ local units, where $d$ is the input dimension. Denote the $i$-th local unit as $\mathcal{X}_i$ and denote $\mathcal{Z}_i=\mathcal{X}_i\times \mathcal{Y}$ as the $i$-th local region of $\mathcal{Z}$, then we have the following proposition \ref{prop.1}:
\newtheorem{proposition}{Proposition}[section]
\newtheorem{theorem}{Theorem}[section]
\newtheorem{assumption}{Assumption}[section]

\begin{proposition}
\label{prop.1}
Under the partition $\{\mathcal{Z}_i\}_{i=1}^{L_m}$ of the sample space $\mathcal{Z}$, the error $\mathcal{R}_{\mathcal{M} \mid \mathcal{S}}$ has the following decomposition:
\begin{equation}
\label{eq:decomposition}
\begin{aligned}
&\mathcal{R}_{\mathcal{M} \mid \mathcal{S}}=\sum_{i=1}^{L_m} \mathbb{P}\left(\boldsymbol{z}\in \mathcal{Z}_i\right) \cdot \mathbb{E}_{\boldsymbol{z}\mid \mathcal{Z}_i}\left[\text{tr}\left[\text{Cov}_{\mathcal{M} \mid \mathcal{S}}\left[f^\sigma\left(\boldsymbol{x}; \boldsymbol{\hat{\theta}}_\mathcal{M}\right)\right]\right]\right]\\
&+\mathbb{E}_{\boldsymbol{z}}\left[\left\|\mathbb{E}_{\mathcal{M} \mid \mathcal{S}}\left[f^\sigma\left(\boldsymbol{x}; \boldsymbol{\hat{\theta}}_\mathcal{M}\right)\right] - f^*\left(\boldsymbol{x}\right) \right\|_2^2 \right],\\
\end{aligned}
\end{equation}
where $\text{tr}(\cdot)$ denotes the trace operator, $\mathbb{E}_{\boldsymbol{z}\mid\mathcal{Z}_i}$ is the conditional expectation of $\boldsymbol{z}$ under the condition $\boldsymbol{z}\in\mathcal{Z}_i$, and $\text{Cov}_{\mathcal{M}\mid\mathcal{S}}$ denotes the conditional covariance matrix of $f^\sigma\left(\boldsymbol{x}; \boldsymbol{\hat{\theta}}_\mathcal{M}\right)$ with respect to $\mathcal{M}$ given $\mathcal{S}$.
\end{proposition}

\begin{figure*}[t]
    \centering
    \includegraphics[width=0.98\textwidth]{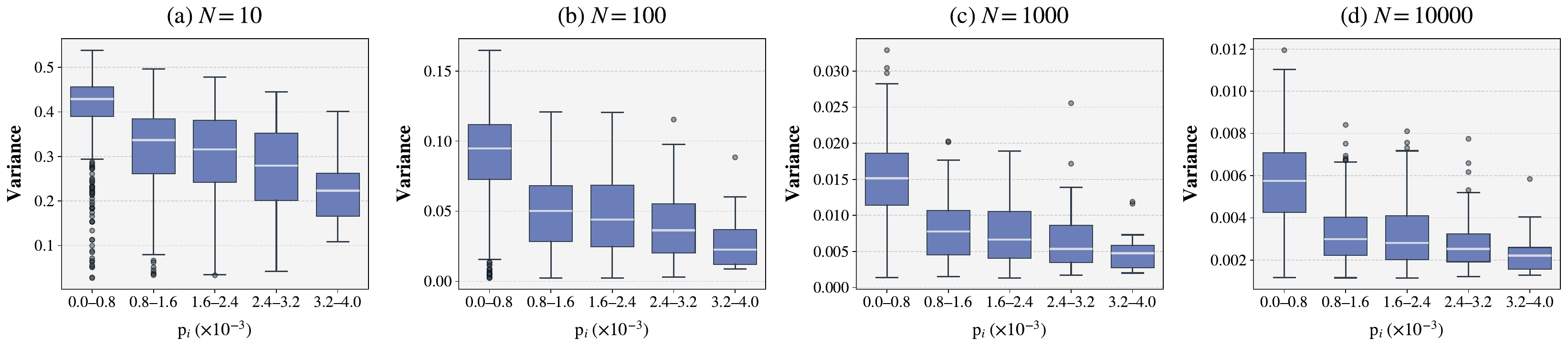}  
    \caption{(a)-(d): Boxplot of the conditional model variance across local regions with different resampling probabilities $\mathrm{p}_i$ (binned) under different memory buffer sizes $N$ ((a) corresponds to $N=10$, (b) to $N=100$, (c) to $N=1000$ and (d) to $N=10000$).}  
    \label{fig:theo_valid}
\end{figure*}

\noindent\textbf{Remark.} In Eq. \eqref{eq:decomposition}, the first term is driven primarily by variance, which can be reduced by appropriately tuning $\pi\left(\boldsymbol{z}\right)$. The second term, on the other hand, is mainly determined by the inherent bias between the model $f^\sigma(\cdot; \boldsymbol{\hat{\theta}}_\mathcal{M})$ and the Bayes-optimal model $f^*(\cdot)$. 
\subsection{Behavior of Variance in Local Regions}
Since the model bias in Eq. \eqref{eq:decomposition} is primarily dominated by the model's intrinsic structural error, while the variance term is directly influenced by the resampling strategy, we proceed to analyze the variance component in further detail to provide insights for designing resampling strategies for different local regions. Before proceeding with the subsequent discussion, we first introduce the following two reasonable assumptions:

\begin{assumption}
\label{assump_1}
For each local region $\mathcal{Z}_i$ and any sample $\left(\bm{x}, y\right) \in \mathcal{Z}_i$, where $1 \le i \le L_m$, it holds that $\left\|f^\sigma\left(\boldsymbol{x}; \boldsymbol{\hat{\theta}}_{\mathcal{M}_i}\right) - f^\sigma\left(\boldsymbol{x}; \boldsymbol{\hat{\theta}}_\mathcal{M}\right)\right\|_2 \le \gamma$, where $\mathcal{M}_i = \mathcal{M} \cap \mathcal{Z}_i$.
\end{assumption}

\begin{assumption}
\label{assump_2}
For any two distinct datasets $\mathcal{S}$ and $\widetilde{\mathcal{S}}$, it holds that for all $\bm{x} \in \mathcal{X}$,  $\left\|f^\sigma\left(\boldsymbol{x}; \boldsymbol{\hat{\theta}}_{\mathcal{S}}\right) - f^\sigma\left(\boldsymbol{x}; \boldsymbol{\hat{\theta}}_{\widetilde{\mathcal{S}}}\right)\right\|_2 \le \phi\left(h\left(\mathcal{S}, \widetilde{\mathcal{S}}\right)\right)$, where $h(\cdot, \cdot)$ denotes the Hausdorff distance \cite{hausdorff1978grundzuge}, and $\phi(\cdot)$ is a non-decreasing bounded function satisfying $0\le u_0\le u < +\infty$, here $u_0=\inf_{t\ge 0}\phi(t)$ and $u=\sup_{t\ge 0}\phi(t)$.
\end{assumption}

Due to the fact that deep learning models typically exhibit local behavior during training \cite{GaoQLGLXZ23}, meaning the model's output within a specific local neighborhood of the sample space is primarily determined by the training samples in that region \cite{shamir2021dimpled}. Therefore, Assumption \ref{assump_1} is reasonable. Meanwhile, Assumption \ref{assump_2} states that ERM models trained on two “closer” datasets should have more similar predictive behaviors. This aligns with the fundamental understanding of data-driven model training: small changes in the training data should not lead to significant changes in model behavior. Moreover, similar properties have already been discussed and validated in prior works \cite{RosenfeldG23,babyadaptive}, further supporting the reasonableness of this assumption.

With the two assumptions stated above, we present the following Theorem \ref{theo.1}:
\begin{theorem}
\label{theo.1}
Under Assumptions \ref{assump_1} and \ref{assump_2}, given any local partition $\{\mathcal{Z}_i\}_{i=1}^{L_m}$ of the sample space $\mathcal{Z}$, and the corresponding index sets $\mathcal{I}_i = \{ j \in \{1, 2, \dots, n\} \mid \boldsymbol{z}_j \in \mathcal{Z}_i \cap \mathcal{S} \}$ of samples in each partitioned region, the following inequality holds for each $\mathcal{Z}_i$:
\begin{equation}
\label{eq:main_theorem}
\begin{aligned}
&\mathbb{E}_{\boldsymbol{z}\mid \mathcal{Z}_i}\text{tr}\left[\text{Cov}_{\mathcal{M} \mid \mathcal{S}}\left[f^\sigma\left(\boldsymbol{x}; \boldsymbol{\hat{\theta}}_\mathcal{M}\right)\right]\right]\\
&\le \mathbbm{1}\left(\mathcal{Z}_i \cap \mathcal{S} \ne \emptyset\right) \bigg[ \left(1-\left(1-\mathrm{p}_{i}\right)^N\right) \Big[C_0+\left(1-\mathrm{p}_{i}\right)^N C_1\Big] \\
& + 2l_iC_2\left[\left(1-\mathrm{p}_{i}l_i^{-1}\right)^N-\left(1-\mathrm{p}_{i}\right)^N\right]\left(1-\left(1-\mathrm{p}_{i}l_i^{-1}\right)^{N}\right)\bigg]\\
&+4\gamma^2,
\end{aligned}
\end{equation}
where $C_0 = \left(u_0 + 2\gamma\right)u_0$, $C_1 = 2(u + 2\gamma)u - C_0$, and $C_2 = \left(\phi(m) + 2\gamma\right)\phi(m) - C_0$. The quantity $\mathrm{p}_i = \frac{\sum_{j \in \mathcal{I}_i} \pi(\boldsymbol{z}_j)}{\sum_{k=1}^n \pi(\boldsymbol{z}_k)}$ is the probability that a sample drawn from $\mathcal{M}$ according to the resampling probablity mass $\pi(\boldsymbol{z})$ belongs to the $i$-th local region $\mathcal{Z}_i$, $l_i=\left|\mathcal{Z}_i\cap\mathcal{S}\right|$ denotes the number of samples from $\mathcal{S}$ that fall into the $i$-th local region.
\end{theorem}

\noindent\textbf{Remark.} Eq. \eqref{eq:main_theorem} provides an upper bound on the conditional model variance, expressed in terms of the secondary sampling probabilities $\mathrm{p}_i$ for each local region. For all $i \in \Omega_{-\emptyset}$, the terms $\left(1 - \left(1 - \mathrm{p}_{i}\right)^N\right)$ and $\left(1 - \left(1 - \mathrm{p}_{i}l_i^{-1}\right)^N\right)$ in Eq \eqref{eq:main_theorem} rapidly approach $1$ as the sampling budget $N$ of the Memory Buffer increases. Consequently, the upper bound in Eq \eqref{eq:main_theorem} becomes dominated by $\left(1 - \mathrm{p}_{i}\right)^N$ and $\left(1 - \mathrm{p}_{i}l_i^{-1}\right)^N$. This implies that a larger $\mathrm{p}_{i}$ promotes faster decay of both $\left(1-\mathrm{p}_{i}\right)^N$ and $\left(1-\mathrm{p}_{i}l_i^{-1}\right)^N$ toward $0$, thereby helping to suppress the local variance in the corresponding region $\mathcal{Z}_i$.

The above statements and theorem lead to the following conclusion:  
(\textbf{C1}) \textbf{For each local region $\mathcal{Z}_i$, a higher resampling probability $\mathrm{p}_i$ helps reduce the model's conditional variance within that region}.

To further verify these two quantitative relationships, we design corresponding validation experiment \textbf{ValEx} to empirically assess their validity. \textbf{ValEx}: we sample a 10-class classification dataset from a mixture of ten two-dimensional Gaussian distributions. We first place $K=10$ angles $\left\{\theta_{i}\right\}_{i=1}^{10}$ uniformly spaced in the interval $[0, 2\pi]$, and use them to generate the mean vectors $\left\{\left(3\cos \theta_{i}, 3\sin \theta_{i}\right)^\top\right\}_{i=1}^{10}$ for the ten Gaussian components, each with a common covariance matrix $2\mathbf{I}$, where $\mathbf{I}$ is the identity matrix. We first draw a class label $y$ uniformly from $k = \{1, 2, \dots, 10\}$, and then sample an input $\bm{x}$ from the $y$-th Gaussian distribution, forming a data sample $\bm{z} = (\bm{x}, y)$. Using this process, we generate 100000 training samples and 100000 test samples for the training and test datasets, respectively. We then define a square region of side length 20 centered at the origin $(0,0)$ as the considered input space. This input space is partitioned into $(10 / 0.4)^2 = 2500$ local units with grid size $m = 0.4$, resulting in 2500 local regions of the sample space. Before training, we select only those local regions that contain samples in both the training and test sets for analysis, ensuring that every examined region has corresponding training and test samples.

For each choice of memory buffer size $N = 10, 100, 1000, 1000$, we conduct 100 independent trials. In each trial, we initialize a 3-layer feed-forward neural network with ReLU activations and a hidden dimension of 64. We then uniformly sample $N$ instances from the training set to form the memory buffer $\mathcal{M}$, and compute the resampling probability $\mathrm{p}_i$ for each examined local region accordingly. The model is then trained on $\mathcal{M}$ for 50 epochs using the SGD optimizer with a learning rate of 0.1 and a cosine annealing learning rate schedule with 10 linear warmup epochs. For all trained models, we evaluate them on the test set to compute the conditional variance within each local region $\mathcal{Z}_i$.

Based on these results, we draw boxplots (binned by dividing the range of resampling probabilities $\mathrm{p}_i$ into five equal intervals) to visualize how the local variance of the model in each region $\mathcal{Z}_i$ varies with the corresponding resampling probability $\mathrm{p}_i$, for each choice of $N$. As shown in Figs \ref{fig:theo_valid} (a)-(d), for every $N$, the local conditional variance consistently decreases as $\mathrm{p}_i$ increases, which validates conclusion \textbf{C1}. 

\subsection{Sample Selection with Data Density Awareness}
\label{DDA}
To provide concrete guidance for the selection of resampling strategies, we further present Corollary \ref{Corollary.1}, which extends the single-region variance upper bound in Theorem \ref{theo.1} to an upper bound on the overall conditional variance.

\newtheorem{corollary}{Corollary}[section] 

\begin{corollary}
\label{Corollary.1}
Under the basic assumptions and conditions of Theorem \ref{theo.1}, the following inequality holds:
\begin{equation}
\label{eq:main_corollary}
\begin{aligned}
&\sum_{i=1}^{L_m}  \mathbb{P}\left(\boldsymbol{z}\in \mathcal{Z}_i\right) \cdot \mathbb{E}_{\boldsymbol{z}\mid \mathcal{Z}_i}\text{tr}\left[\text{Cov}_{\mathcal{M} \mid \mathcal{S}}\left[f^\sigma\left(\boldsymbol{x}; \boldsymbol{\hat{\theta}}_\mathcal{M}\right)\right]\right]\\
&\le\sum_{i\in\Omega_{-\emptyset}}\mathbb{P}\left(\boldsymbol{z}\in \mathcal{Z}_i\right) \cdot \bigg[ \left(1-\left(1-\mathrm{p}_{i}\right)^N\right) \Big[C_0+\left(1-\mathrm{p}_{i}\right)^N C_1\Big] \\
& + 2l_iC_2\left[\left(1-\mathrm{p}_{i}l_i^{-1}\right)^N-\left(1-\mathrm{p}_{i}\right)^N\right]\left(1-\left(1-\mathrm{p}_{i}l_i^{-1}\right)^{N}\right)\bigg]\\&+4\gamma^2,
\end{aligned}
\end{equation}
where $\Omega_{-\emptyset} = \left\{ i \in \{1, 2, \dots, L_m\} \mid \mathcal{Z}_i \cap \mathcal{S} \ne \emptyset \right\}$ is the index set of all local regions $\mathcal{Z}_i$ such that $\mathcal{Z}_i \cap \mathcal{S} \ne \emptyset$.
\end{corollary}

\noindent\textbf{Remark.} Corollary \ref{Corollary.1} further indicates that the local conditional variance in regions with larger $\mathbb{P}\left(\boldsymbol{z}\in \mathcal{Z}_i\right)$ has a more dominant influence on the overall variance upper bound. Therefore, prioritizing higher resampling probabilities $\mathrm{p}_i$ for regions with larger $\mathbb{P}\left(\boldsymbol{z}\in \mathcal{Z}_i\right)$ is more effective in suppressing the overall variance upper bound.

In summary, the above analysis leads to conclusion (\textbf{C2}): \textbf{To help reduce the conditional MSE $\mathcal{R}_{\mathcal{M} \mid \mathcal{S}}$, resampling should prioritize high-probability regions of the sample distribution, i.e., the resampling probability mass $\pi(\boldsymbol{z})$ should primarily focus on regions where $p(\boldsymbol{z})$ is large}.

\begin{figure}[t]
    \centering
    \includegraphics[width=0.48\textwidth]{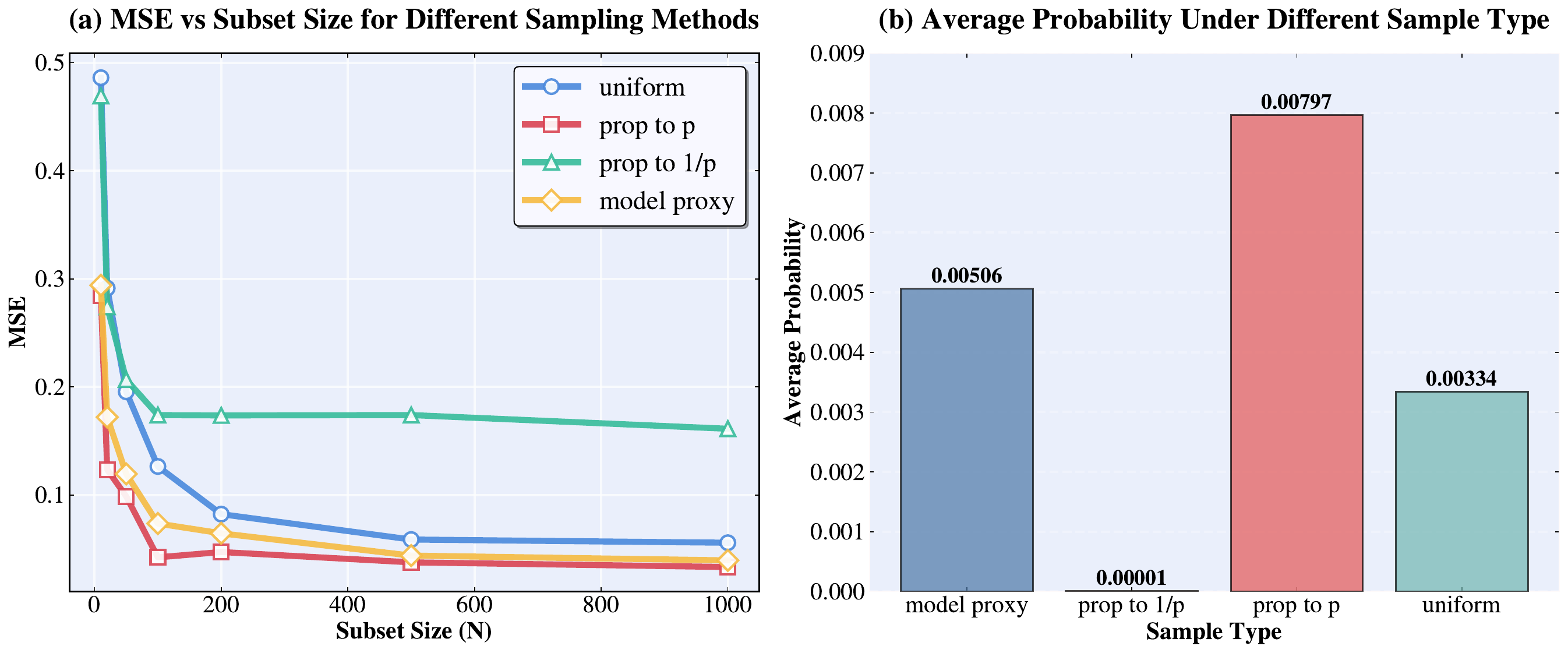}  
    \caption{(a): Curves of $\mathcal{R}_{\mathcal{M} \mid \mathcal{S}}$ under different selection strategies across various values of $N$; (b): Average density of the regions containing the samples in $\mathcal{M}$ under different selection strategies.}  
    \vspace{-3pt}
    \label{fig:sample_type_compare}
    \vspace{-5pt}
\end{figure}

To verify conclusion \textbf{C2}, based on the setup of the validation experiment \textbf{ValEx}, we compute the curves of conditional MSE $\mathcal{R}_{\mathcal{M} \mathcal{S}}$ under different $N$ selections across three sampling strategies: \ding{172} Resampling $\mathcal{M}$ from $\mathcal{S}$ with uniform probability, i.e., $\pi(\bm{z}_i)=\frac{1}{n}$ for $\forall 1\le i\le n$; \ding{173} Resampling according to the probability mass allocation $\pi(\bm{z}_i) = \frac{p(\bm{z}_i)}{\sum_{j=1}^{n} p(\bm{z}_j)}$, which prioritizes samples with higher probability density; and \ding{174} For each sample $\bm{z}_i$, resampling according to $\pi(\bm{z}_i) = \frac{1/p(\bm{z}_i)}{\sum_{j=1}^{n}(1/p(\bm{z}_j))}$, which prioritizes samples with lower probability density, serving as a control to strategy \ding{173}. Since the real data distributions are known, the Bayes-optimal classifier can be explicitly constructed as $f^*(\boldsymbol{x})_k = \frac{p(\boldsymbol{x} \mid y=k) \cdot p(y=k)}{\sum_{j=1}^{10} p(\boldsymbol{x} \mid y=j) \cdot p(y=j)}$. Then we plot in Fig. \ref{fig:sample_type_compare} (a) the average $\mathcal{R}_{\mathcal{M} \mid \mathcal{S}}$ over 10 independent trials for models trained on $\mathcal{M}$ selected under each strategy, for $N = 10, 20, 50, 100, 200, 500, 1000, 2000, 5000$. Compared with uniform sampling, favouring samples with high density (refered as ``prop to p") yields significantly lower MSE across all $N$, whereas preferring low-density samples (refered as ``prop to 1/p") produces markedly higher MSE. These results validates conclusion \textbf{C2}. 

The analysis above indicates that prioritizing high-density samples can enhance the quality of selected samples. However, the density $p(\bm{z})$ is typically unknown in practice, which precludes the direct implementation of density-guided selection strategies like \ding{173}. To address this challenge, in the next section, we present an adaptive density estimation method based on learned feature representations to guide sample selection.

\section{Density-Aware Sample Selection for CL}
In this section, we propose a density-aware sample selection method for coreset construction. First, since the sample density $p(\bm{z})$ is unknown, we introduce PGM to estimate densities in the feature space. Specifically, we first perform VMP to alleviate the density sparsity problem caused by the curse of dimensionality, and then use the GMM in the low-dimensional space to model the sample density. Based on this estimation method, we propose PDAC, which constructs the coreset by selecting high-density samples. Subsequently, we propose SPDAC, which incrementally updates the PGM parameters via the streaming EM algorithm to adapt to streaming scenarios.

% \vspace{-10pt}
\subsection{Estimating Sample Density via PGM}
Owing to the powerful representation capacity of deep learning models, their feature space efficiently encodes the semantic structure and distributional patterns of the raw data, making density modeling feasible \cite{LafonRRT23,QiCGLLWZ24}. However, the curse of dimensionality renders high-dimensional features extremely sparse \cite{HastieTF09}, which hinders direct density estimation. To address this challenge, we first apply VMP \cite{abdi2010principal} to project the features onto a low-dimensional subspace while preserving the principal modes of variation in the data distribution.

Let $\bm{h} = f_{\psi}\left(\bm{x};\bm{\theta}\right)$ denote the feature representation of input sample $\bm{x}$ extracted by the backbone network, where $f_\psi\left(\cdot\,;\bm{\theta}\right):\mathcal{X}\rightarrow\mathbb{R}^{D}$ outputs the activation value of the last layer before the classification head, and $D$ is the dimension of the feature space. Let $\mathcal{C}_i^t$ denote the index set of samples from class $i$ in $\mathcal{S}^t$, defined as $\mathcal{C}_{i}^t=\left\{1\le j\le\left|\mathcal{S}^t\right| \mid y_{t,j}=i, \left(\bm{x}_{t,j}, y_{t,j}\right)\in\mathcal{S}^t \right\}$. For each task $t$ and each class $y \in \mathcal{Y}^t$, we compute the mean feature vector as $\bar{\bm{h}}_{t}^y = \frac{1}{\left|\mathcal{C}_y^t\right|} \sum_{i \in \mathcal{C}_y^t} \bm{h}_{t,i}$, and construct the centered feature covariance matrix $\bar{\bm{H}}_y=\frac{1}{\left|\mathcal{C}_y^t\right|} \sum_{i \in \mathcal{C}_y^t} (\bm{h}_{t,i} - \bar{\bm{h}}_{t}^y)(\bm{h}_{t,i} - \bar{\bm{h}}_{t}^y)^\top \in \mathbb{R}^{D \times D}$. Subsequently, we perform spectral decomposition on $\bar{\bm{H}}_y$ to obtain $\bar{\bm{H}}_y=\bm{V}_{y}\bm{S}_{y}\bm{V}_{y}^{-1}$, where the columns of $\bm{V}_y \in \mathbb{R}^{D \times D}$ are the eigenvectors ordered by descending eigenvalues. Taking the first $d$ eigenvectors $\bm{v}_1, \dots, \bm{v}_d$ constitutes the VMP matrix corresponding to class $y$:
\begin{equation}
    \bm{W}_y^t = \left[\bm{v}_1, \dots, \bm{v}_d\right] \in \mathbb{R}^{D \times d}.
\end{equation}

Using the VMP matrix, we compute the projected feature for each sample as $\bm{\xi}_{t,i} = (\bm{W}_y^t)^\top \left( \bm{h}_{t,i} - \bar{\bm{h}}_t^y \right) \in \mathbb{R}^d$ which preserves the principal directions of variation within class $y$ while reducing the feature dimensionality from $D$ to $d \ll D$, thereby helping to mitigate the issue of high-dimensional sparsity. Based on this, we construct a GMM \cite{bishop2006pattern} for each class $y\in\mathcal{Y}^{t}$ to model the intra-class conditional density of the projected features:
\begin{equation}
    p\left(\bm{\xi}\mid y\right)=\sum_{l=1}^L \alpha_{y}^l \mathcal{N}\left( \bm{\xi}\Big|\bm{\mu}_{y}^l, \bm{\Sigma}_{y}^l \right),
\end{equation}
where $L$ is the number of Gaussian components, $\mathcal{N}\left( \bm{\xi} \,\Big| \, \bm{\mu}_{y}^l, \bm{\Sigma}_{y}^l \right)$ is the probability density function of the $l$-th Gaussian component, and $\alpha_{y}^l>0$, $\bm{\mu}_{y}^l\in\mathbb{R}^d$, $\bm{\Sigma}_{y}^l\in\mathbb{R}^{d\times d}$ denote the corresponding mixing weight, mean vector, and covariance matrix, respectively. These parameters are estimated by maximizing the following log-likelihood: 
\begin{equation}
\mathcal{L}\left(\alpha_{y}^l,\bm{\mu}_{y}^l,\bm{\Sigma}_{y}^l\right)=\sum_{i \in \mathcal{C}_y^t}\log p\left(\bm{\xi}_{t,i}\mid y\right),
\end{equation}
which is optimized using the EM algorithm. Specifically, we perform iterative estimation by alternately executing the \textbf{E-step} and \textbf{M-step} as described below:

\noindent \textbf{E-step}: The posterior responsibility of each projected feature $\bm{\xi}_{t,i}$ ($i\in\mathcal{C}_{y}^t$) for each of its corresponding Gaussian components is calculated as follows:
\begin{equation}
\label{eq:e-step}
r_{t,i}^l = \frac{ \alpha_{y}^l \, \mathcal{N}\left( \bm{\xi}_{t,i} \mid \bm{\mu}_{y}^l, \bm{\Sigma}_{y}^l \right) }{ p(\bm{\xi}_{t,i} \mid y) }, \quad 1\le l\le L.
\end{equation}
\noindent\textbf{M-step}: Update the corresponding parameters of each Gaussian component:
\begin{equation}
\label{eq:m-step}
\begin{aligned}
    \alpha_y^l &= \frac{1}{|\mathcal{C}_y^t|} \sum_{i \in \mathcal{C}_y^t} r_{t,i}^l, \quad 
    \bm{\mu}_{y}^l = \frac{ \sum_{i \in \mathcal{C}_y^t} r_{t,i}^l \bm{\xi}_{t,i} }{ \sum_{i \in \mathcal{C}_y^t} r_{t,i}^l }, \\
    \bm{\Sigma}_{y}^l &= \frac{ \sum_{i \in \mathcal{C}_y^t} r_{t,i}^l (\bm{\xi}_{t,i} - \bm{\mu}_{y}^l)(\bm{\xi}_{t,i} - \bm{\mu}_{y}^l)^\top }{ \sum_{i \in \mathcal{C}_y^t} r_{t,i}^l }.
\end{aligned}
\end{equation}
The final GMM parameters are obtained by iterating the aforementioned E-step and M-step for at most $G$ times. Then the VMP and GMM processes described above collectively form the final PGM pipeline. Subsequently, we use the calculated conditional density $p\left(\bm{\xi}_{t,i}\mid y\right)$ as a proxy for the conditional density of $\bm{x}_{t,i}$, denoted as $\hat{p}\left(\bm{x}_{t,i} \mid y\right) \triangleq p(\bm{\xi}_{t,i} \mid y)$ ($i\in\mathcal{C}_{y}^t$). Furthermore, we use the empirical frequency to estimate the probability of class $y$: $\hat{p}(y) = \frac{n_y}{\sum_{y'} n_{y'}}$, where $n_y$ is the number of observed samples belonging to class $y$. Ultimately, the overall density estimate for the sample $\bm{z}_{t,i}=\left(\bm{x}_{t,i},y\right)$ is formulated as $\hat{p}\left(\bm{z}_{t,i}\right)=\hat{p}\left(\bm{x}_{t,i}\mid y\right)\cdot \hat{p}(y)$, which is utilized for sample selection.

To verify the feasibility of this estimation scheme, we incorporate sampling strategy \ding{175} into the setup of \textbf{ValEX}, where $\mathcal{M}$ is sampled according to the probability $\pi\left(\bm{z}_i\right)=\frac{\hat{p}\left(\bm{z}_i\right)}{\sum_{j=1}^n\hat{p}\left(\bm{z}_j\right)}$. This strategy is then compared with strategies \ding{172}, \ding{173}, and \ding{174} presented in \ref{DDA}. Here we set the number of components in PGM to $L=3$, the projection dimension to $d=5$, and the number of EM iterations to $G=20$. For initialization, the means of each PGM components are initialized by randomly selecting projected features, and the covariance matrices are initialized as identity matrices. We then plot the mean curve of conditional MSE under different choice of $N$ across 10 independent experiments in Fig. \ref{fig:sample_type_compare}(a) (corresponding to ``model proxy''). The result demonstrates that, compared with Uniform sampling, the strategy based on estimated density priority effectively reduces the conditional MSE for all values of $N$, and exhibits behavior close to that of the strategy directly based on real density priority (``prop to p''). Additionally, we measure the mean of the real average density of samples in $\mathcal{M}$ by each strategy under various $N$, which is recorded in the bar chart shown in Fig. \ref{fig:sample_type_compare}(b). In contrast to Uniform sampling, the strategy based on estimated density priority significantly enhances the density of the sampled samples. The results above further confirm the effectiveness of the proposed estimated density in approximating the real density and its reliability in sample selection.
\subsection{Buffer Selection with PDAC}
\begin{algorithm}[t]
\caption{Procedure of PDAC during the training of $\mathcal{T}^t$}
\label{algo:1}
\begin{algorithmic}[1]
\REQUIRE
    Current task $\mathcal{T}^t$, training set $\mathcal{S}^t$, memory buffer $\mathcal{M}$, task buffer size $\{N_{t,j}\}_{j \leq t}$.
\ENSURE Updated replay buffer $\mathcal{M}$.
\FOR{$\forall y\in\mathcal{Y}^t$}
\STATE Initialize the PGM component corresponding to class $y$ using the samples in $\mathcal{S}^t$ that satisfy $i \in \mathcal{C}_i^t$, and perform $G$ iterations of EM updates based on Eqs \eqref{eq:e-step} and \eqref{eq:m-step}.
\ENDFOR
\STATE Compute $\hat{p}(\bm{z}) \gets \hat{p}(\bm{x} \mid y) \cdot \hat{p}(y)$ for $\forall\bm{z}\in\mathcal{S}^t$
\STATE Compute probability $\pi(\bm{z}) \gets \dfrac{\hat{p}(\bm{z})}{\sum_{\bm{z}' \in \mathcal{S}^t} \hat{p}(\bm{z}')}$ for $\forall\bm{z}\in\mathcal{S}^t$
\STATE Select $N_{t,t}$ samples from $\mathcal{S}^t$ by multinomial sampling according to $\pi(\bm{z})$ and add to $\mathcal{M}$.
\IF{$\left|\mathcal{M}\right|>N$}
\FOR{each previous task $\mathcal{T}^{t'}$ with $t' < t$}
    \STATE Compute $\hat{p}(\bm{z}) \gets \hat{p}(\bm{x} \mid y) \cdot \hat{p}(y)$ for $\forall \bm{z}\in\mathcal{S}^{t'}\cap\mathcal{M}$
    \STATE For each $\bm{z} \in \mathcal{E}^{t'}$, compute $w(\bm{z}) \gets 1 - \hat{p}(\bm{z})$
    \STATE Compute $\pi'(\bm{z}) \gets \dfrac{1-\hat{p}(\bm{z})}{\sum_{\bm{z}' \in \mathcal{S}^{t'}\cap\mathcal{M}} \left(1-\hat{p}(\bm{z}')\right)}$ for $\forall\bm{z}\in\mathcal{S}^{t'}\cap\mathcal{M}$
    \STATE Select $N_{t-1,t'}-N_{t,t'}$ samples from $\mathcal{S}^{t'}\cap\mathcal{M}$ by multinomial sampling according to $\pi'(\bm{z})$ and remove them from $\mathcal{M}$.
\ENDFOR
\ENDIF
\end{algorithmic}
\end{algorithm}
Based on the estimated sample density $\hat{p}(\bm{z}) = \hat{p}(\bm{x} \mid y) \cdot \hat{p}(y)$ for $\bm{z}\in\mathcal{Z}$ obtained from the PGM, we propose PDAC (formalized in Algorithm \ref{algo:1}) to guide the construction of the core set. Specifically, for each task $\mathcal{T}^t$, PDAC first computes the estimated density $\hat{p}(\bm{z}_{t,i})$ for each sample $\bm{z}_{t,i}=\left(\bm{x}_{t,i},t_{t,i}\right)$ in the current training set $\mathcal{S}^t$. Subsequently, it computes the resampling probability $\pi\left(\bm{z}_{t,i}\right)=\frac{\hat{p}(\bm{z}_{t,i})}{\sum_{\bm{z}\in\mathcal{S}^t} \hat{p}(\bm{z})}$ and performs multinomial sampling to select $N_{t,t}$ samples to add to $\mathcal{M}$, where $N_{i,j}$ denotes the buffer size allocated to samples from $\mathcal{S}^j$ in $t$-th task's training. When the buffer is full, for each preceding task $\mathcal{T}^{t'}$ ($t'< t$), if its current stored sample count $N_{t-1,t'}\ge N_{t,t'}$ (the sample size for $\mathcal{T}^{t'}$ in the training phase of $\mathcal{T}^{t-1}$) exceeds the pre-allocated capacity $N_{t,t'}$, then multinomial sampling is performed according to the probability $\pi\left(\bm{z}_{t-1,i}\right)=\frac{1-\hat{p}(\bm{z}_{t-1,i})}{\sum_{\bm{z}\in\mathcal{S}^{t-1}}\left(1 - \hat{p}(\bm{z})\right)}$ to select $N_{t-1,t'}-N_{t,t'}$ samples as candidates for replacement, thereby ensuring that samples with high estimated densities are prioritized for retention.
\subsection{Adapting to Streaming Data with SPDAC}
\begin{algorithm}[t]
\caption{Procedure of SPDAC for an incoming batch $\mathcal{B}$}
\label{algo:2}
\begin{algorithmic}[1]
\REQUIRE
Current batch $\mathcal{B} = \{(\bm{x}_i, y_i)\}$,
Memory buffer $\mathcal{M}$,
task buffer sizes $\{N_{t,j}\}_{j \leq t}$, update step size $\beta \in (0,1)$.
\ENSURE Updated buffer $\mathcal{M}$ and PGM parameters.
\FOR{each $\bm{z}=(\bm{x},y)\in\mathcal{B}$}
\STATE Update $n_y=n_y^{\text{old}}+ \left|\mathcal{B}_{y}\right|$
\IF{PGM for $y$ is not initialized}
\STATE Perform initialization of PGM parameters for $y$
\ELSE
\STATE Update VMP and PGM parameters of class $y$ according to Eq. \eqref{eq:proj_update}, \eqref{eq:e-step-streaming} and \eqref{eq:m-step-streaming}
\ENDIF
\ENDFOR

\STATE Compute $\hat{p}(\bm{z}) \gets \hat{p}(\bm{\xi} \mid y) \cdot \hat{p}(y)$ for each $\bm{z} = (\bm{x}, y) \in \mathcal{B}$
\STATE Compute $\pi(\bm{z}) \gets \dfrac{\hat{p}(\bm{z})}{\sum_{\bm{z}' \in \mathcal{B}} \hat{p}(\bm{z}')}$ for each $\bm{z} \in \mathcal{B}$
\FOR{each $\bm{z} \in \mathcal{B}$}
\IF{$ \left|\mathcal{M}\right| < N$}
\STATE Add $\bm{z}$ to $\mathcal{M}$
\ELSE
\STATE Sample $c \sim \text{Uniform}\{1, 2, \dots, n\}$
\STATE Compute $\epsilon(\bm{z}) \gets \pi(\bm{z}) \cdot \mathcal{B} $
\IF{$c \cdot \epsilon(\bm{z}) \le N$}
\STATE Sample $i \sim \text{Uniform}\{1, 2, \dots, M\}$
\STATE Replace the $i$-th sample in $\mathcal{M}$ with $\mathbf{z}$.
\ENDIF
\ENDIF
\ENDFOR
\end{algorithmic}
\end{algorithm}
The parameters of PDAC are updated based on all observed samples of the current task. To further extend PDAC to streaming scenarios where samples arrive in batches, we propose SPDAC for streaming data (as described in Algorithm \ref{algo:2}). Specifically, for the current batch $\mathcal{B}$, we first check its label set $\mathcal{Y}_{\mathcal{B}} = \{ y \mid (\bm{x}, y) \in \mathcal{B} \}$. For each class $y\in\mathcal{Y}_{\mathcal{B}}$, we update its corresponding sample count $n_y=n_y^{\text{old}}+ \left|\mathcal{B}_{y}\right|$, where $\mathcal{B}_{y}=\left\{(\bm{x}, y') \in \mathcal{B} : y' = y\right\}$, where $n_{\text{old}}$ denotes the old count before update. Subsequently, for each class whose corresponding PGM parameters have not been initialized, we perform initialization using the samples in $\mathcal{B}$. Then for each already initialized class $y\in\mathcal{Y}_{\mathcal{B}}$, we incrementally update its feature mean $\bar{\bm{h}}^y$ and the corresponding centered feature covariance matrix $\bar{\bm{H}}^y$:
\begin{equation}
\label{eq:proj_update}
\begin{aligned}
\bar{\bm{h}}^{y} &= \frac{\bar{\bm{h}}^{y,\,\text{old}} \cdot n_y^{\text{old}} + \sum_{\bm{z}\in\mathcal{B}_y}\bm{h}}{n_y^{\text{old}}+\left|\mathcal{B}_y\right|},\\
\bar{\bm{H}}^{y} &= \frac{\bar{\bm{H}}^{y,\,\text{old}}\cdot n_y^{\text{old}} + \sum_{\bm{z}\in\mathcal{B}_y}(\bm{h}-\bar{\bm{h}}^{y})(\bm{h}-\bar{\bm{h}}^{y})^\top}{n_y^{\text{old}}+\left|\mathcal{B}_y\right|}.
\end{aligned}
\end{equation}

Next, for each Gaussian component parameter in the PGM corresponding to each class $y$, we perform EM updates as shown in Eq. \eqref{eq:e-step} and \eqref{eq:m-step} using the data in $\mathcal{B}$. We then update the mean and covariance parameters of the Gaussian components based on EMA, forming the streaming EM process as follows:

\noindent\textbf{E-step} (streaming): Calculate the responsibility for each Gaussian component based on the projected features $\bm{\xi}$ corresponding to each sample $\bm{z}=\left(\bm{x},y\right)\in\mathcal{B}_{y}$:
\begin{equation}
\label{eq:e-step-streaming}
r^l = \frac{ \alpha_{y}^l \, \mathcal{N}\left( \bm{\xi} \mid \bm{\mu}_{y}^l, \bm{\Sigma}_{y}^l \right) }{ p(\bm{\xi} \mid y) }, \quad 1\le l\le L.
\end{equation}
\noindent\textbf{M-step} (streaming): Update the corresponding parameters of each Gaussian component based on EMA:
\begin{equation}
\label{eq:m-step-streaming}
\begin{aligned}
    \alpha_y^l &= \frac{1}{|\mathcal{B}_y|} \sum_{\bm{z} \in \mathcal{B}_y} r^l,\, 
    \bm{\mu}_y^{l} = \left(1-\beta\right)\bm{\mu}_y^{l,\,\text{old}}+\beta\frac{ \sum_{\bm{z} \in \mathcal{B}_y} r^l \bm{\xi} }{ \sum_{\bm{z} \in \mathcal{B}_y} r^l }, \\
    \bm{\Sigma}_{y}^{l} &= (1-\beta)\bm{\Sigma}_{y}^{l,\,\text{old}} + \beta\frac{ \sum_{\bm{z}\in \mathcal{B}_y} r^l (\bm{\xi} - \bm{\mu}_y^{l})(\bm{\xi} - \bm{\mu}_y^{l})^\top }{ \sum_{\bm{z}\in \mathcal{B}_y} r^l },
\end{aligned}
\end{equation}
where $\beta$ denotes the update step size. For the current batch $\mathcal{B}$, we calculate the estimated density $\hat{p}(\bm{z})$ for each sample $\bm{z} \in \mathcal{B}$, as well as the resampling probability $\pi(\bm{z})=\frac{\hat{p}(\bm{z})}{\sum_{\bm{z}'\in\mathcal{B}}\hat{p}(\bm{z}')}$. Subsequently, similar to the reservoir sampling technique \cite{vitter1985random}, for $\forall\bm{z}\in\mathcal{B}$, we first randomly sample a positive integer $c$ from $[1, n]$, where $n$ is the number of all samples observed so far. Then using the resampling probability, we construct a scaling factor $\epsilon(\bm{z})=\pi(\bm{z})\cdot\left|\mathcal{B}\right|$. If $c\cdot\epsilon(\bm{z})\le N$, the sample $\bm{z}$ is added to $\mathcal{M}$, and if $\left|\mathcal{M}\right|>N$, a sample within $\mathcal{M}$ is randomly replaced. By executing the SPDAC process introduced above for each batch, the PGM parameters can be promoted to adapt to streaming data. Assisted by its estimated density, SPDAC enables the dynamic construction of high-quality memory buffer in streaming scenarios.

\section{Experiments}
In this section, to thoroughly evaluate our proposed method, we conduct comparative experiments under two different CL settings: offline and streaming. 
% \vspace{-10pt}

\subsection{Experimental Setup}
\label{sec:setup}
\subsubsection{Datasets}
Following the common practice in CL research \cite{TiwariKIS22,HaoJL23}, we conduct experiments on the following three CL datasets, which are widely used in other works:

\noindent\textbf{Split-CIFAR10}: The CIFAR10 dataset \cite{krizhevsky2009learning} consists of $10$ classes, with $5000$ training samples and $1000$ testing samples per class. We split it into $T=5$ tasks, where each task $\mathcal{T}^t$ contains $\left|\mathcal{Y}^t\right|=2$ classes, forming the Split-CIFAR10 dataset.

\noindent\textbf{Split-CIFAR100}: The CIFAR100 dataset \cite{krizhevsky2009learning} comprises $100$ classes, , with $500$ training samples and $100$ test samples per class. We split it into $T=10$ tasks, where each task $\mathcal{T}^t$ contains $\left|\mathcal{Y}^t\right|=10$ classes, forming Split-CIFAR100.

\noindent\textbf{Split-TinyImageNet}: The TinyImageNet dataset \cite{wu2017tiny} consists of $200$ classes, with $500$ training samples and $50$ test samples per class. We divide it into $T=10$ tasks, where each task includes $\left|\mathcal{Y}^t\right|=20$ classes, yielding the Split-TinyImageNet.

\subsubsection{Evaluation Metrics}
Following the work of \cite{BorsosM020,ShimMJSKJ21}, we consider evaluation under the class-incremental setting, where the task identifier is not provided during testing, which is both more realistic and challenging. Under this evaluation setting, we adopt the following two metrics to assess the algorithm's performance:

\begin{table*}[htbp]
\centering
\caption{Results of comparative experiments on Split-CIFAR10 and Split-CIFAR100 under the offline CL setting. The mean values and corresponding standard deviations of three independent experiments are reported. The best results are displayed in bold, while the second-best results are $\underline{\text{underlined}}$. ``$\uparrow$'' and ``$\downarrow$'' indicate "the higher the better" and "the lower the better", respectively.}

\label{tab:1}
\resizebox{0.95\textwidth}{!}{%
\begin{tabular}{lcccccccc}
      \toprule
      \multirow{4}{*}{\textbf{Method}} & \multicolumn{4}{c}{\textbf{Balanced}} & \multicolumn{4}{c}{\textbf{Imbalanced}} \\
      \cmidrule(lr){2-5} \cmidrule(lr){6-9}
      & \multicolumn{2}{c}{$N=200$} & \multicolumn{2}{c}{$N=500$} & \multicolumn{2}{c}{$N=200$} & \multicolumn{2}{c}{$N=500$} \\
      \cmidrule(lr){2-3} \cmidrule(lr){4-5} \cmidrule(lr){6-7} \cmidrule(lr){8-9}
      & ACC ($\uparrow$)& FM ($\downarrow$)& ACC ($\uparrow$) & FM ($\downarrow$)& ACC ($\uparrow$) & FM ($\downarrow$)& ACC ($\uparrow$) & FM ($\downarrow$)\\
\hline
\rowcolor{softblue}\multicolumn{9}{c}{\textbf{Split-CIFAR10}} \\
\hline
Uniform Sampling & 48.26$\pm$0.55 & 55.86$\pm$0.91 & 59.10$\pm$1.75 & 41.58$\pm$1.71 & 47.15$\pm$1.84 & 56.63$\pm$2.34 & 57.47$\pm$1.52 & 42.96$\pm$1.76 \\
k-Means of Features \cite{NguyenLBT18} & 47.96$\pm$1.82 & 56.18$\pm$2.39 & 58.22$\pm$2.60 & 42.81$\pm$1.80 & 46.32$\pm$2.85 & 57.30$\pm$1.25 & 57.76$\pm$1.28 & 42.78$\pm$1.19\\
k-Center of Embeddings \cite{SenerS18} & 47.55$\pm$3.80 & 55.70$\pm$1.41 & 54.97$\pm$2.99 & 41.88$\pm$1.86 & 45.96$\pm$3.20 & 56.12$\pm$0.99 & 56.63$\pm$1.95 & 42.92$\pm$1.65\\
Hardest Samples \cite{AljundiKT19} & 39.43$\pm$3.10 & 56.50$\pm$2.73 & 48.84$\pm$1.56 & 45.98$\pm$2.62 & 36.20$\pm$1.29 & 60.36$\pm$2.40 & 46.55$\pm$1.26 & 45.44$\pm$0.59\\
Highest Prob & 40.94$\pm$1.96 & 56.67$\pm$3.31 & 47.51$\pm$1.43 & 44.69$\pm$1.99 & 37.82$\pm$0.81 & 58.24$\pm$2.82 & 48.03$\pm$1.72 & 45.21$\pm$1.52 \\
iCaRL' Selection \cite{RebuffiKSL17} & 47.26$\pm$3.35 & 56.42$\pm$1.41 & 58.57$\pm$1.61 & 42.22$\pm$1.49 & 47.65$\pm$0.97 & 56.92$\pm$1.52 & 57.70$\pm$1.83 & 43.04$\pm$1.70\\
Greedy Coreset \cite{BorsosM020} & 48.98$\pm$1.24 & 56.05$\pm$1.55 & 60.31$\pm$1.63 & 41.03$\pm$1.45 & 47.88$\pm$1.09 & 55.94$\pm$1.84 & 58.33$\pm$1.08 & 42.49$\pm$1.36\\
OCS \cite{YoonMYH22} & 48.13$\pm$1.20 & 56.62$\pm$1.44 & 59.22$\pm$2.05 & 41.30$\pm$1.91 & 46.85$\pm$1.21 & 56.91$\pm$1.38 & 57.12$\pm$1.92 & 44.13$\pm$1.42\\
GCR \cite{TiwariKIS22} & 48.67$\pm$1.33 & 55.98$\pm$0.96 & 59.65$\pm$1.46 & 42.04$\pm$1.50 & 47.51$\pm$1.05 & 56.22$\pm$1.51 & 58.17$\pm$1.34 & 43.08$\pm$1.27\\
PBCS \cite{ZhouPZLCZ22} & 50.78$\pm$0.75 & 54.25$\pm$1.08 & \underline{61.41$\pm$1.65} & \underline{40.08$\pm$1.69} & 49.93$\pm$1.82 & 55.36$\pm$0.85 & \underline{60.15$\pm$1.83} & \underline{40.07$\pm$1.65}\\
BCSR \cite{HaoJL23} & \underline{51.04$\pm$0.92} & \underline{53.79$\pm$0.98} & 61.17$\pm$1.39 & 40.35$\pm$1.72 & \underline{50.21$\pm$1.31} & \underline{54.89$\pm$1.26} & 59.86$\pm$1.45 & 41.12$\pm$1.58\\
\rowcolor{softgray}PDAC (Ours) & \textbf{52.56$\pm$0.81} & \textbf{53.08$\pm$1.25} & \textbf{64.70$\pm$1.52} & \textbf{39.20$\pm$1.25} & \textbf{51.22$\pm$0.47} & \textbf{53.44$\pm$0.90} & \textbf{63.09$\pm$1.44} & \textbf{38.85$\pm$1.90}\\
\hline
\rowcolor{softblue}\multicolumn{9}{c}{\textbf{Split-CIFAR100}} \\
\hline
Uniform Sampling & 13.30$\pm$0.59 & 72.33$\pm$0.91 & 19.45$\pm$0.95 & 65.15$\pm$0.53 & 12.65$\pm$0.63 & 73.54$\pm$1.33 & 19.21$\pm$0.69 & 65.51$\pm$0.65 \\
k-Means of Features \cite{NguyenLBT18} & 13.51$\pm$1.02 & 72.31$\pm$1.82 & 19.52$\pm$0.87 & 64.93$\pm$0.97 & 12.91$\pm$0.93 & 73.05$\pm$1.06 & 19.53$\pm$0.94 & 64.45$\pm$0.98\\
k-Center of Embeddings \cite{SenerS18} & 13.48$\pm$0.92 & 73.77$\pm$1.14 & 19.42$\pm$0.45 & 65.09$\pm$0.32 & 13.05$\pm$0.54 & 73.89$\pm$0.74 & 19.16$\pm$0.62 & 64.34$\pm$0.54\\
Hardest Samples \cite{AljundiKT19} & 10.58$\pm$0.46 & 76.89$\pm$0.72 & 14.53$\pm$0.56 & 69.99$\pm$1.35 & 11.38$\pm$0.57 & 75.37$\pm$1.12 & 13.49$\pm$0.66 & 70.44$\pm$0.21\\
Highest Prob & 11.30$\pm$0.71 & 76.65$\pm$1.32 & 14.05$\pm$0.48 & 72.86$\pm$1.97 & 11.19$\pm$0.54 & 76.93$\pm$1.11 & 13.36$\pm$0.93 & 71.24$\pm$0.37\\
iCaRL' Selection \cite{RebuffiKSL17} & 13.86$\pm$0.86 & 72.67$\pm$0.92 & 20.08$\pm$0.49 & 64.97$\pm$0.94 & 13.65$\pm$0.66 & 74.09$\pm$0.69 & 19.68$\pm$0.42 & 64.72$\pm$0.42\\
Greedy Coreset \cite{BorsosM020} & 14.18$\pm$0.77 & 71.75$\pm$0.84 & 20.19$\pm$0.69 & 64.65$\pm$0.57 & 14.02$\pm$0.48 & 73.12$\pm$0.98 & 20.21$\pm$0.73 & 63.51$\pm$0.67\\
OCS \cite{YoonMYH22} & 13.10$\pm$0.56 & 73.15$\pm$0.40 & 19.31$\pm$0.52 & 65.09$\pm$0.71 & 12.27$\pm$0.76 & 73.29$\pm$0.80 & 19.32$\pm$0.65 & 64.98$\pm$0.73\\
GCR \cite{TiwariKIS22} & 13.35$\pm$0.49 & 72.66$\pm$0.83 & 19.26$\pm$0.68 & 64.82$\pm$0.78 & 12.53$\pm$0.60 & 73.76$\pm$0.92 & 19.09$\pm$0.45 & 65.52$\pm$0.83\\
PBCS \cite{ZhouPZLCZ22} & 14.14$\pm$0.78 & \underline{71.94$\pm$0.97} & 20.47$\pm$0.66 & \underline{63.74$\pm$0.80} & \underline{14.05$\pm$0.27} & 73.10$\pm$0.93 & \underline{20.77$\pm$0.56} & \underline{63.47$\pm$0.52}\\
BCSR \cite{HaoJL23} & \underline{14.25$\pm$0.81} & 72.20$\pm$0.68 & \underline{20.44$\pm$0.79} & 64.17$\pm$0.61 & 13.89$\pm$0.41 & \underline{72.84$\pm$0.85} & 20.65$\pm$0.64 & 63.51$\pm$0.83\\
\rowcolor{softgray}PDAC (Ours) & \textbf{15.18$\pm$0.42} & \textbf{71.10$\pm$1.16} & \textbf{21.86$\pm$0.64} & \textbf{62.70$\pm$0.85} & \textbf{14.95$\pm$0.37} & \textbf{72.35$\pm$0.73} & \textbf{21.41$\pm$0.66} & \textbf{62.28$\pm$1.00}\\
\hline
\end{tabular}%
}
\vspace{-5pt}

\end{table*}

\noindent\textbf{Average ACcuracy (ACC)}: ACC is defined as $\text{ACC} = \frac{1}{T}\sum_{t=1}^T A_{T,t}$, where $A_{i,j}$ denotes the accuracy of the model on task $\mathcal{T}^j$ after the model has been trained on task $\mathcal{T}^i$.

\noindent\textbf{Forgetting Metric (FM)}: FM is defined as the average of the maximum performance drops on all previous tasks after the model has been trained on all $T$ tasks, formulated as $\mathrm{FM} = \frac{1}{T-1}\sum_{i=1}^{T-1} \max_{1\le j\le T-1} (A_{j,i} - A_{T,i})$.

\begin{table}[t]
\centering
\caption{The average sample selection time (in seconds) of different coreset selection methods across three independent runs on the Split-CIFAR10 and Split-CIFAR100 datasets under the Offline CL setting, with buffer sizes $N=200$ and $N=500$.}
\label{tab:2}
\resizebox{0.48\textwidth}{!}{%
\begin{tabular}{lcccc}
      \toprule
      \multirow{2}{*}{\textbf{Method}} & \multicolumn{2}{c}{\textbf{Split-CIFAR10}} & \multicolumn{2}{c}{\textbf{Split-CIFAR100}} \\
      \cmidrule(lr){2-3} \cmidrule(lr){4-5} 
      & $N=200$ & $N=500$ & $N=200$ & $N=500$ \\
\hline
Uniform Sampling & 0.21 & 0.51 & 0.30 & 0.71\\
k-Means of Features \cite{NguyenLBT18} & 33.80 & 32.62 & 36.16 & 37.24\\
k-Center of Embeddings \cite{SenerS18} & 28.74 & 36.85 & 31.46 & 36.32\\
Hardest Samples \cite{AljundiKT19} & 22.13 & 22.26 & 26.65 & 27.80\\
Highest Prob & 22.15 & 22.32 & 26.87 & 27.32\\
iCaRL' Selection \cite{RebuffiKSL17} & 22.18 & 22.27 & 27.57 & 27.54\\
Greedy Coreset \cite{BorsosM020} & 363.10 & 1130.85 & 967.99 & 3750.93\\
OCS \cite{YoonMYH22} & 204.35 & 639.28 & 542.65 & 2102.07\\
GCR \cite{TiwariKIS22} & 61.43 & 152.49 & 146.43 & 562.95\\
PBCS \cite{ZhouPZLCZ22} & 88.79 & 207.35 & 186.88 & 701.25\\
BCSR \cite{HaoJL23} & 181.95 & 488.30 & 385.12 & 1537.49 \\
\rowcolor{softgray}PDAC (Ours) & 23.95 & 24.04 & 36.42 & 37.67\\
\hline
\end{tabular}%
}
\vspace{-5pt}

\end{table}

\subsubsection{Model Architecture And Training Settings}
Following the widespread practice in CL research \cite{BorsosM020, TiwariKIS22}, we adopt ResNet18 \cite{HeZRS16} as the backbone architecture of the learning model across all datasets. For model training, we consistently employ the SGD optimizer without momentum or weight decay in all experiments, with a constant learning rate of $ 0.03 $ and a training/replay batch size of $ 32 $. For the offline setting, we uniformly train the learning model for $10$ epochs in all experiments, and for the streaming CL setting, we use two training epoch settings, including $10$ epochs and $1$ epoch, to comprehensively evaluate the algorithm's adaptability on streaming data. Additionally, we use consistent data augmentation across all datasets, which involves first padding the image by $4$ pixels, followed by randomly cropping to its original size, and performing a random horizontal flip. All experiments are run on a server equipped with a Tesla V100-SXM2 16GB GPU. For all experiments, we run three independent trials and record the mean and standard deviation to report the error bars.

\begin{figure}[t]
    \centering
    \includegraphics[width=0.48\textwidth]{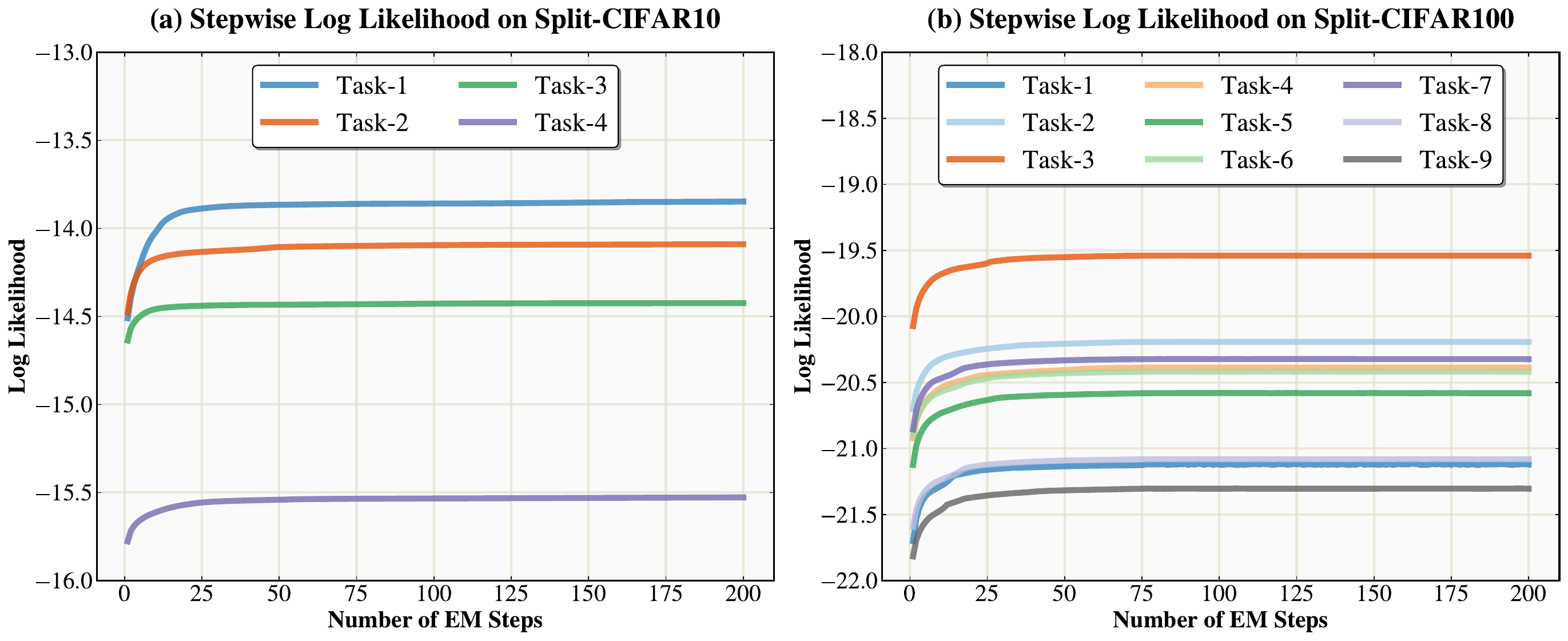}
    \caption{Stepwise sample log-likelihood of the PGM component's EM iterations across different tasks. Panels (a) and (b) show the results on Split-CIFAR10 and Split-CIFAR100, respectively.}  
    \vspace{-3pt}
    \label{fig:3}
    \vspace{-5pt}
\end{figure}

\subsection{Offline Continual Learning}
We first evaluate PDAC under the offline CL setting. In this setting, the training set of the current task is accessible at any time during the training of the current task and only becomes inaccessible when the training of the next task starts. Selection is performed after the training of each task $ \mathcal{T}^t $ and before the start of the next task. During the training of task $\mathcal{T}^t$, the buffer size corresponding to each task $ \mathcal{T}^i $ is uniformly divided as $N_{t,i}=\frac{N}{t}$. We conduct evaluations on Split-CIFAR10 and Split-CIFAR100 with two different buffer sizes: $ N=200 $ and $ N=500 $. Following the work of \cite{ZhouPZLCZ22}, we evaluate all methods under both balanced and imbalanced scenarios. The balanced scenario means that the number of samples for all classes in each task is consistent. For the imbalanced scenario, we follow the approach in \cite{CuiJLSB19} to adjust the number of samples $ n_y $ for each class $ y $ in the training set of each task using the formula $ n_{y}^{\text{imb}} = 0.05^{y\,\bmod \frac{|\mathcal{Y}|}{T}} \cdot n_y $, where $ y\,\bmod \frac{|\mathcal{Y}|}{T} $ denotes the index of class $ y $ within the current task. Under the aforementioned setting, we compare PDAC with existing CS methods to purely compare the performance and efficiency of PDAC against current CS strategies in terms of sample selection.

\begin{table*}[htbp]
\centering
\caption{Performance comparison on Split-CIFAR10, Split-CIFAR100, and Split-TinyImageNet under two streaming CL configurations: Relaxed Streaming (10 epochs) and Online Streaming (1 epoch). Results show the mean and standard deviation over three independent runs, with the best and second-best marked in bold and $\underline{\text{underlined}}$, respectively.}
\label{tab:3}
\resizebox{0.9\textwidth}{!}{%
\begin{tabular}{lcccccccc}
      \toprule
      \multirow{4}{*}{\textbf{Method}}  & \multicolumn{4}{c}{\textbf{Relaxed Streaming}} & \multicolumn{4}{c}{\textbf{Online Streaming}} \\
      & \multicolumn{2}{c}{$N=200$} & \multicolumn{2}{c}{$N=500$} & \multicolumn{2}{c}{$N=200$} & \multicolumn{2}{c}{$N=500$} \\
      \cmidrule(lr){2-3} \cmidrule(lr){4-5} \cmidrule(lr){6-7} \cmidrule(lr){8-9}
      & ACC ($\uparrow$)& FM ($\downarrow$)& ACC ($\uparrow$) & FM ($\downarrow$)& ACC ($\uparrow$) & FM ($\downarrow$)& ACC ($\uparrow$) & FM ($\downarrow$) \\
\hline
\rowcolor{softblue}\multicolumn{9}{c}{\textbf{Split-CIFAR10}}\\
\hline
ER \cite{chaudhry2019tiny} & 49.38$\pm$0.48 & 57.17$\pm$0.72 & 58.85$\pm$2.37 & 44.35$\pm$3.58 & 40.48$\pm$0.82 & 47.25$\pm$1.72 & 49.70$\pm$1.76 & 35.11$\pm$2.17\\
ER-MIR \cite{aljundi2019online} & 49.92$\pm$0.94 & 57.05$\pm$1.25 & 59.38$\pm$2.30 & 44.01$\pm$2.96 & 40.41$\pm$0.93 & 47.23$\pm$1.08 & 48.38$\pm$1.24 & 35.72$\pm$1.98\\
GSS \cite{AljundiLGB19} & 44.25$\pm$3.87 & 63.10$\pm$4.54 & 59.12$\pm$2.21 & 44.15$\pm$3.34 & 39.38$\pm$0.45 & 48.08$\pm$2.12 & 43.71$\pm$2.62 & 38.26$\pm$4.74\\
ASER \cite{ShimMJSKJ21} & 48.89$\pm$2.10 & 57.43$\pm$1.93 & 59.25$\pm$1.89 & 43.39$\pm$1.49 & 40.39$\pm$0.92 & 47.39$\pm$2.32 & 48.69$\pm$1.58 & 35.49$\pm$2.05\\
MetaSP \cite{SunLS0W22} & \underline{51.49$\pm$2.15} & \underline{55.65$\pm$2.01} & \underline{60.98$\pm$1.47} & \underline{42.41$\pm$1.32} & \underline{41.87$\pm$1.30} & \underline{44.72$\pm$1.98} & \underline{50.33$\pm$1.47} & \underline{34.07$\pm$1.64}\\
\rowcolor{softgray}SPDAC (Ours) & \textbf{55.05$\pm$1.08} & \textbf{48.02$\pm$1.07} & \textbf{63.82$\pm$1.01} & \textbf{36.60$\pm$0.82} & \textbf{44.95$\pm$1.12} & \textbf{40.41$\pm$2.19} & \textbf{51.67$\pm$2.15} & \textbf{30.49$\pm$2.32}\\
\hline
\rowcolor{softblue}\multicolumn{9}{c}{\textbf{Split-CIFAR100}}\\
\hline
ER \cite{chaudhry2019tiny} & 13.09$\pm$0.83 & 75.88$\pm$0.49 & 19.24$\pm$0.78 & 68.99$\pm$1.38 & 10.78$\pm$0.37 & 47.05$\pm$1.09 & 13.63 $\pm$ 0.60 & 37.61$\pm$0.81\\
ER-MIR \cite{aljundi2019online} & 13.15$\pm$0.97 & 75.43$\pm$0.82 & 19.18$\pm$0.99 & 68.74$\pm$0.85 & 10.50$\pm$0.83 & 47.43$\pm$0.59 & 13.25$\pm$0.71 & 37.57$\pm$0.76\\
GSS \cite{AljundiLGB19} & 11.24$\pm$0.50 & 76.26$\pm$0.95 & 19.54$\pm$0.66 & 68.50$\pm$1.18 & 8.94$\pm$0.34 & 48.63$\pm$0.57 & 10.72$\pm$0.42 & 38.82$\pm$0.94\\
ASER \cite{ShimMJSKJ21} & 13.89$\pm$0.91 & 75.78$\pm$1.30 & 19.82$\pm$1.01 & \underline{67.12$\pm$0.83} & 10.35$\pm$0.84 & 48.01$\pm$0.83 & 12.97$\pm$0.86 & 38.04$\pm$0.87\\
MetaSP \cite{SunLS0W22} & \underline{14.39$\pm$0.48} & \underline{74.80$\pm$0.76} & \underline{20.21$\pm$0.80} & 67.32$\pm$0.57 & \underline{10.85$\pm$0.40} & \underline{46.71$\pm$0.54} & \underline{13.98$\pm$0.93} & \underline{37.55$\pm$0.60}\\
\rowcolor{softgray}SPDAC (Ours) & \textbf{15.09$\pm$0.38} & \textbf{73.86$\pm$0.48} & \textbf{21.69$\pm$0.81} & \textbf{65.99$\pm$0.84} & \textbf{12.22$\pm$0.72} & \textbf{44.49$\pm$1.26} & \textbf{15.32$\pm$0.45} & \textbf{37.02$\pm$0.54}\\
\hline
\rowcolor{softblue}\multicolumn{9}{c}{\textbf{Split-TinyImageNet}}\\
\hline
ER \cite{chaudhry2019tiny} & 7.79$\pm$0.65 & 67.27$\pm$0.74 & 9.38$\pm$0.48 & 65.96$\pm$0.55 & \underline{5.90$\pm$0.28} & \underline{43.67$\pm$0.52} & \underline{8.51$\pm$0.44} & 40.70$\pm$0.55\\
ER-MIR \cite{aljundi2019online} & 7.68$\pm$0.79 & 67.66$\pm$1.07 & 9.01$\pm$0.93 & 66.13$\pm$0.98 & 5.87$\pm$0.31 & 43.75$\pm$0.48 & 7.91$\pm$0.50 & 40.54$\pm$0.72\\
GSS \cite{AljundiLGB19} & 7.56$\pm$0.45 & 67.92$\pm$0.83 & 9.31$\pm$0.43 & 65.84$\pm$0.78 & 5.85$\pm$0.12 & 43.88$\pm$1.35 & 7.04$\pm$0.44 & 41.97$\pm$0.81\\
ASER \cite{ShimMJSKJ21} & 7.92$\pm$0.34 & \underline{67.01$\pm$0.50} & 9.40$\pm$0.31 & 65.77$\pm$0.67 & 5.73$\pm$0.30 & 43.80$\pm$0.69 & 7.82$\pm$0.57 & \underline{40.15$\pm$0.37}\\
MetaSP \cite{SunLS0W22} & \underline{8.14$\pm$0.39} & 67.03$\pm$0.44 & \underline{9.93$\pm$0.52} & \underline{65.19$\pm$0.45} & 5.84$\pm$0.45 & 43.69$\pm$0.39 & 8.48$\pm$0.46 & 40.49$\pm$0.38\\
\rowcolor{softgray}SPDAC (Ours) & \textbf{8.87$\pm$0.41} & \textbf{66.57$\pm$0.67} & \textbf{10.85$\pm$0.39} & \textbf{64.79$\pm$0.46} & \textbf{6.81$\pm$0.22} & \textbf{42.56$\pm$0.73} & \textbf{9.16$\pm$0.56} & \textbf{38.90$\pm$0.54}\\
\hline
\end{tabular}%
}
\end{table*}

Under the aforementioned setting, we compare PDAC with existing CS methods to purely compare the performance and efficiency of our scheme against current CS strategies in terms of sample selection. The compared methods include Uniform Sampling, k-Means of Features \cite{NguyenLBT18}, k-Center of Embeddings \cite{SenerS18}, retaining the hardest-to-classify points (Hardest Samples) \cite{AljundiKT19}, the selection strategy of iCaRL \cite{RebuffiKSL17}, Greedy Coreset \cite{BorsosM020}, OCS \cite{YoonMYH22}, the selection strategy of GCR \cite{TiwariKIS22}, PBCS \cite{ZhouPZLCZ22}, and BCSR \cite{HaoJL23}. As a control, we also introduce a strategy that prioritizes the selection of samples with high model output confidence (Highest Prob). For other methods, we adopt their original hyperparameter settings, and for baselines that provide sample importance scores (excluding Uniform Sampling), we also employ an importance-prioritized buffer sample replacement strategy similar to that in Algorithm \ref{algo:1}. For PDAC, we select a projection dimension of $ d=10 $, a number of Gaussian components of $ L=7 $, and a number of EM iterations of $ G=20 $.

\begin{table}[t]
\centering
\caption{Average runtime (in minutes) of various rehearsal-based streaming CL methods across three independent experiments on the Split-CIFAR10, Split-CIFAR100, and Split-TinyImageNet datasets under the relaxed streaming configuration.}
\label{tab:4}
\resizebox{0.48\textwidth}{!}{%
\begin{tabular}{lcccccc}
      \toprule
      \multirow{2}{*}{\textbf{Method}} & \multicolumn{2}{c}{\textbf{Split-CIFAR10}} & \multicolumn{2}{c}{\textbf{Split-CIFAR100}} & \multicolumn{2}{c}{\textbf{Split-TinyImageNet}} \\
      \cmidrule(lr){2-3} \cmidrule(lr){4-5} \cmidrule(lr){6-7}
      & $N=200$ & $N=500$ & $N=200$ & $N=500$ & $N=200$ & $N=500$\\
\hline
ER \cite{chaudhry2019tiny} & 32.36 & 34.63 & 36.64 & 38.72 & 79.96 & 82.91\\
ER-MIR \cite{aljundi2019online} & 43.84 & 46.97 & 48.05 & 48.56 & 105.39 & 113.48\\
GSS \cite{AljundiLGB19} & 98.52 & 121.51 & 108.32 & 129.40 & 255.74 & 267.32\\
ASER \cite{ShimMJSKJ21} & 77.53 & 113.35 & 83.45 & 109.77 & 185.47 & 204.10\\
MetaSP \cite{SunLS0W22} & 193.05 & 203.45 & 198.85 & 210.29 & 546.60 & 564.13\\
\rowcolor{softgray}SPDAC (Ours) & 43.68 & 44.49 & 96.82 & 97.24 & 285.17 & 285.47\\
\hline
\end{tabular}%
}
\end{table}

The results are presented in Tab. \ref{tab:1}. Across both the Split-CIFAR10 and Split-CIFAR100 datasets, PDAC outperforms all other baseline CS methods in terms of both ACC and FM metrics, under both balanced and imbalanced scenarios. Specifically, on Split-CIFAR10 with a buffer size of $ N=500 $, PDAC achieves an ACC of $ 64.70\% $ in the balanced setting, which is $ 3.29\% $ higher than that of the second-best method. Meanwhile, the forgetting rate decreases to $39.20\%$, which is $0.88\%$ lower than the second-best method. On Split-CIFAR100 with $ N=200 $, PDAC’s ACC exceeds the second-best method by $ 0.93\% $, while its forgetting rate is $ 0.84\% $ lower than that of the second-best method. Notably, PDAC maintains its leading advantage even in the imbalanced scenario, which further validates its robustness against class distribution shifts.

Furthermore, we report the average selection time (in seconds) of each method across three independent experiments under different buffer sizes $ N $ on each dataset in Tab. \ref{tab:2}. Owing to the reduced density computation cost afforded by PGM's projection dimensionality reduction, PDAC exhibits almost no disadvantage in selection time compared to other non-purely random selection strategies. Notably, compared to other methods involving bilevel optimization, such as Greedy Coreset, BCSR, and PBCS, PDAC's operational efficiency is several, or even tens of times faster. Furthermore, since PDAC's selection importance is evaluated at the sample level and does not require repeated leave-one-out validation on coreset candidates at the set level, its selection time is independent of the buffer size $ N $. This makes PDAC’s efficiency advantage more pronounced when the buffer size $ N $ is larger.

\subsection{Streaming Continual Learning}
In the streaming CL setting, samples arrive in batches. Once the next batch begins training, the previous batch becomes inaccessible, requiring the algorithm to decide which samples to select and replace during the current batch's learning process. Under this setting, we evaluate the performance of the SPDAC method on streaming data using the Split-CIFAR10, Split-CIFAR100, and Split-TinyImageNet datasets with buffer sizes $N=200$ and $N=500$. We compare SPDAC with existing rehearsal-based streaming CL methods that rely on data selection, including ER \cite{chaudhry2019tiny} based on reservoir sampling, GSS \cite{AljundiLGB19} based on gradient diversity, ER-MIR \cite{aljundi2019online} with maximum interference retrieval, ASER \cite{ShimMJSKJ21} based on adversarial shapley value, and MetaSP \cite{SunLS0W22} based on example influence. For the other baselines, we follow their original hyperparameter settings. For SPDAC, we select the EMA update step size $\beta=0.5$ and set the number of EM algorithm update steps $G=1$. Here we consider two distinct training epoch configurations as follows:

\noindent\textbf{Relaxed Streaming Configuration}: In this configuration, we adopt a 10 epoch training for each task. However, the selection and the buffer' update strictly follow the streaming constraint, meaning selection can only be performed batch-wise.

\noindent\textbf{Online Streaming Configuration}: In this configuration, the model is trained only 1 epoch per task. Each batch of data is immediately discarded after forward inference and parameter updates, preventing its reuse, which closely resembles training scenarios under resource constraints.

\begin{figure*}[htbp]
    \centering
    \includegraphics[width=0.98\textwidth]{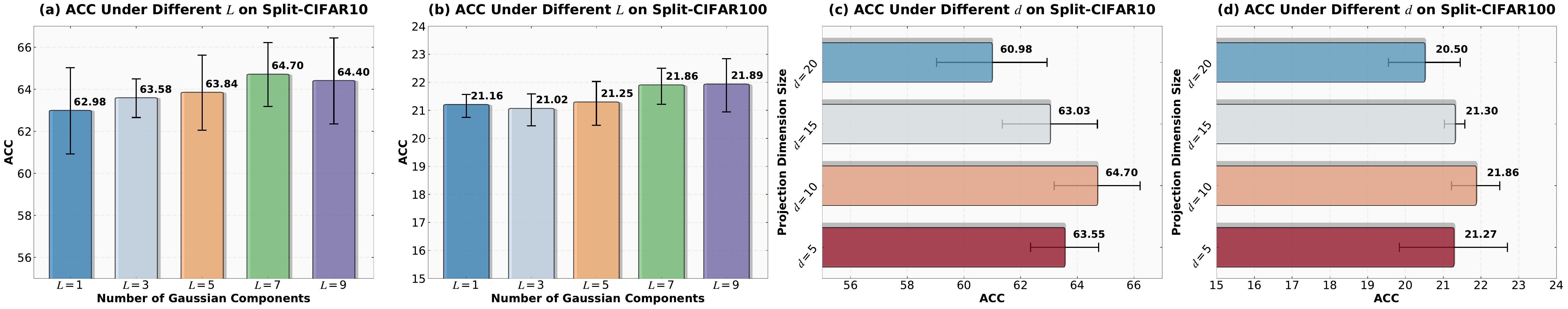}
    \caption{(a)-(b): ACC of PDAC on Split-CIFAR10 and Split-CIFAR100 with varying numbers of Gaussian components $L$, respectively; (c)-(d): ACC on Split-CIFAR10 and Split-CIFAR100 versus the projection dimensions $d$. Error bars represent the standard deviation across three independent runs.}  
    \vspace{-3pt}
    \label{fig:4}
    \vspace{-5pt}
\end{figure*}

The corresponding experimental results are recorded in Tab. \ref{tab:3}. Under both the relaxed streaming and online streaming CL settings, SPDAC consistently surpasses other baselines across all three datasets: Split-CIFAR10, Split-CIFAR100, and Split-TinyImageNet. Specifically, for the Relaxed Streaming case, SPDAC achieves an ACC of $63.82\%$ on the Split-CIFAR10 dataset when $N=500$, outperforming the second-best method by $2.84\%$, while reducing the FM to $36.60\%$, which is $5.81\%$ lower than the second-best method. On the more challenging Split-CIFAR100 and Split-TinyImageNet, SPDAC similarly maintains its leading performance and lowest FM. Notably, SPDAC still delivers outstanding performance even under the more stringent Online Streaming Configuration, which fully demonstrates its strong adaptability in streaming CL. Furthermore, in Tab. \ref{tab:4}, we report the average runtime (in minutes) of each method on each dataset across three independent runs under the relaxed streaming configuration. The results further demonstrate that SPDAC still maintains a relatively favorable runtime efficiency compared to other baselines. Combined with its consistently optimal performance presented in Tab. \ref{tab:3}, this indicates that SPDAC can achieve a desirable balance between performance and efficiency.

\subsection{Ablation Studies}
To analyze how different hyperparameter choices affect the behavioral characteristics of the various components within PDAC and SPDAC, we conduct a series of ablation studies. Unless otherwise specified, we consistently choose a buffer size $N=500$ and 10 training epochs, with the remaining training configurations consistent with those described in \ref{sec:setup}.

\subsubsection{Convergence Efficiency of the EM Algorithm in PGM}

We first analyze the convergence behavior of the EM update process of the PGM component in PDAC with respect to the number of iterations. To this end, we record the average log-likelihood of all training samples for the current task during the EM iterations of PGM on each task across different datasets under the balanced scenario of the offline CL setting, where the number of update steps $G$ ranges from 0 to 200. The other hyperparameters are set to $L=7$ and $d=10$, and the buffer size is $N=500$. The results shown in Fig. \ref{fig:3} demonstrate that for every task on each dataset, the average log likelihood rapidly increases within the first 10$–$15 steps and stabilizes after approximately 20 steps, with further increases in the number of iterations yielding only marginal improvements. This indicates that the EM update of PGM can converge efficiently without requiring a large number of iterations, thereby ensuring the high efficiency of the PDAC algorithm.

% 高斯组件个数$L$对PDAC性能的影响。
\subsubsection{Impact of the Number of Gaussian Components $L$ on PDAC's Performance}
\begin{table}[htbp]
\centering
\caption{Average selection time (in seconds) of PDAC on Split-CIFAR10 and Split-CIFAR100 with different $L$, calculated from three independent experiments.}
\label{tab:5}
\resizebox{0.48\textwidth}{!}{%
\begin{tabular}{lccccc}
      \toprule
      \textbf{Dataset} & $L=1$ & $L=3$ & $L=5$ & $L=7$ & $L=9$ \\
\hline
\textbf{Split-CIFAR10} &  23.32 & 23.47 & 23.76 & 24.04 & 24.32\\
\textbf{Split-CIFAR100} &  34.67 & 35.97 & 36.39 & 37.67 & 39.10\\
\hline
\end{tabular}%
}
\vspace{-5pt}
\end{table}

To fully understand the impact of the number of Gaussian components selected for PGM on PDAC’s performance, we evaluate the ACC and selection time of PDAC with different Gaussian component number settings $L\in\left\{1, 3, 5, 7, 9\right\}$ of PGM under the balanced senario of the offline CL setting. Here, other hyperparameters are set to $G=20$ and $d=10$, and the buffer size is set to $N=500$. As shown in Fig. \ref{fig:4} (a) and (b), the results of ACC demonstrate that as $L$ increases, PDAC’s performance exhibits a consistent upward trend. When $L$ becomes excessively large, the performance gain tends to be marginal and even starts to yield slight negative gains. This phenomenon aligns with the trade-off between model complexity and generalization ability in density estimation \cite{scott2015multivariate}: an appropriate $L$ helps capture the structure of task distributions, while excessively high complexity contributes little to performance improvement. Furthermore, the comparison of selection time in Tab. \ref{tab:5} indicates that, benefiting from the projected dimensionality reduction of PGM, the increase in $L$ has only a minor impact on the selection time. Considering both modeling capacity and efficiency, we adopt $L=7$ as the default configuration for our comparative experiments.

\begin{figure*}[htbp]
    \centering
    \includegraphics[width=0.98\textwidth]{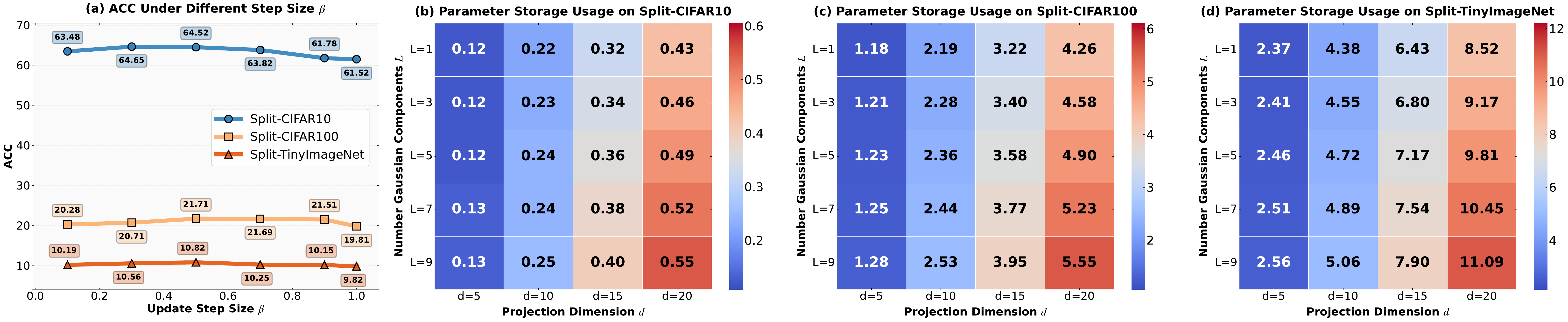}
    \caption{(a): Average ACC of SPDAC across three independent runs on each datasets with different streaming EM update step sizes $\beta \in \left\{0.1, 0.3, 0.5, 0.7, 0.9, 1.0\right\}$ under the streaming CL setting; }  
    \vspace{-3pt}
    \label{fig:5}
    \vspace{-5pt}
\end{figure*}

\begin{table}[htbp]
\centering
\caption{Average selection time (in seconds) of PDAC on Split-CIFAR10 and Split-CIFAR100 with different $d$, calculated from three independent experiments.}
\label{tab:6}
\resizebox{0.43\textwidth}{!}{%
\begin{tabular}{lcccc}
      \toprule
      \textbf{Dataset} & $d=5$ & $d=10$ & $d=15$ & $d=20$\\
\hline
\textbf{Split-CIFAR10} &  23.84 & 24.04 & 26.08 & 26.94\\
\textbf{Split-CIFAR100} &  35.09 & 37.67 & 40.12 & 43.54\\
\hline
\end{tabular}%
}
\end{table}

\subsubsection{The Effect of The Projection Dimension $d$ on PDAC's Performance}
To investigate the impact of the projection dimension $d$ of PGM on PDAC's performance, we evaluate the ACC and selection time of PDAC using different PGM projection dimension settings $d\in\left\{5, 10, 15, 20\right\}$ under the balanced scenario of the offline CL setting. Here, the other parameters are set to $G=20$ and $L=7$, with the buffer size set to $N=500$. The ACC results shown in Fig. \ref{fig:4}(c) and (d) demonstrate that PDAC maintains stable and high performance on both datasets within the range of $d=5$ to $d=15$. However, an excessively high projection dimension inherently exacerbates the issue of feature sparsity in high-dimensional context, which in turn compromises density estimation. Consequently, PDAC's ACC does not show further improvement when $d$ increases to $20$, but instead exhibits a noticeable drop. This indicates that a moderate projection dimension is sufficient to preserve the data's discriminative information for effective density estimation. Considering both stability and efficiency, we adopt $d=10$ as the default configuration for our comparative experiments.

\subsubsection{The Influence of The Update Step Size $\beta$ on SPDAC's Performance}
We further analyze the influence of the update step size $\beta$ of the streaming EM algorithm on SPDAC's performance. Under the streaming CL scenario, we evaluate the average ACC of SPDAC on various datasets across three independent runs, considering different update step size choices $\beta \in \left\{0.1, 0.3, 0.5, 0.7, 0.9, 1.0\right\}$. The remaining hyperparameters are fixed as $L=7$, $d=10$, and the buffer size is set to $N=500$. Since a small $\beta$ (e.g., $\beta=0.1$) causes lagging updates in PGM parameters, making it difficult to adapt to the current data distribution promptly, and an excessively large $\beta$ (e.g., $\beta=0.9$ or $\beta=1.0$) causes the model to respond too quickly to new data information, leading to insufficient retention of historical knowledge, the performance of SPDAC exhibits a trend of initially increasing and then decreasing as $\beta$ increases. A moderate selection of $\beta$ between $0.3$ and $0.7$ results in relatively stable and favorable performance across all datasets, indicating that this range achieves a good balance between fast adaptation and stable memorization. Considering this, we adopt $\beta=0.5$ as the default configuration for our comparative experiments. Moreover, note that when $\beta=1.0$, the EM update degrades into the regular EM update shown in Eqs \eqref{eq:e-step} and \eqref{eq:m-step}. The ACC in this case shows a significant drop on all three datasets. This comparative result clearly demonstrates the effectiveness of the streaming EM algorithm specifically designed for the streaming scenario.

\subsubsection{Storage Usage Analysis of PGM Components}
We quantitatively analyze the storage usage of the PGM component in PDAC. Specifically, we calculate the storage usage (in MB) of the PGM component parameters (including the projection matrix parameters and GMM parameters) under different settings of the two parameters that affect PDAC’s storage usage: projection dimension $d \in \{5, 10, 15, 20\}$ and number of Gaussian components $L \in \{1, 3, 5, 7, 9\}$. As shown in Fig. \ref{fig:5} (b)-(d), the results demonstrate that the storage usage is mainly determined by the projection dimension $ d $. When $ d $ increases, the storage usage of PGM increases almost linearly with $ d $, and this usage is primarily introduced by the projection matrix. Benefiting from the fact that the dimension $ d $ after projection dimensionality reduction is much smaller than the original feature dimension $D$, increasing the number of Gaussian components $ L $ introduces almost no additional storage usage. For example, when $ d=5 $, increasing $ L $ from 1 to 9 on Split-CIFAR10 only brings an additional storage usage of 0.01 MB, while on Split-TinyImageNet, increasing $ L $ from 1 to 9 only results in an additional storage usage of 0.19 MB. This indicates that appropriately limiting the projection dimension can effectively control memory costs while maintaining the modeling capability of PGM. Therefore, we select $ L=7 $ and $ d=10 $ in the comparative experiments.

\section{Conclusion And Future Works}
In this paper, we focus on the balance between efficiency and quality in memory buffer construction for CL problem. Addressing the bottleneck where existing CS methods widely rely on bilevel optimization and incur significant computational overhead, we aim to design a more direct and efficient CS mechanism for memory buffer construction. From the perspective of localized error decomposition, we theoretically reveal the dominant role of samples with high probability density in suppressing the conditional MSE between the buffer-trained model and the Bayes optimal predictor. Inspired by this, we propose the PDAC method, which efficiently estimates the joint density of samples using the PGM model, thereby enabling efficient density-prioritized sampling for buffer construction. Furthermore, we introduce the streaming EM algorithm to enhance the adaptability of PGM parameters to streaming data, extending PDAC to SPDAC for buffer sample selection in streaming scenarios. Extensive experiments demonstrate that the proposed methods outperform other baselines under various CL settings while ensuring favorable efficiency.

Our work provides a new paradigm for efficient and theoretically grounded memory buffer construction in CL problem. In the future, we will explore integrating external density estimators (e.g., generative estimators) to improve density modeling accuracy. Meanwhile, we will further lightweight the design of the density estimation component, making it even more suitable for edge CL scenarios where storage resources are extremely constrained.

\bibliographystyle{IEEEtran}
\bibliography{IEEEabrv,references}

\clearpage
{\appendix
\section*{Proof of proposition \ref{prop.1}}
\label{appendix-1}
\textit{Proof of Proposition \ref{prop.1}:}
Following a decomposition pattern similar to the bias–variance decomposition in \cite{neal2019modern}, we obtain:
\begin{equation*}
\begin{aligned}
\mathcal{R}_{\mathcal{M} \mid \mathcal{S}} &= \mathbb{E}_{\boldsymbol{z}} \mathbb{E}_{\mathcal{M} \mid \mathcal{S}}\left[ \left\| f^\sigma\left(\boldsymbol{x}; \boldsymbol{\hat{\theta}}_\mathcal{M}\right)-\mathbb{E}_{\mathcal{M} \mid \mathcal{S}}\left[f^\sigma\left(\boldsymbol{x}; \boldsymbol{\hat{\theta}}_\mathcal{M}\right)\right]\right\|_2^2\right] \\
&+2\mathbb{E}_{\boldsymbol{z}}\Bigg[\underbrace{\mathbb{E}_{\mathcal{M} \mid \mathcal{S}}\left[f^\sigma\left(\boldsymbol{x}; \boldsymbol{\hat{\theta}}_\mathcal{M}\right)-\mathbb{E}_{\mathcal{M} \mid \mathcal{S}}\left[f^\sigma\left(\boldsymbol{x}; \boldsymbol{\hat{\theta}}_\mathcal{M}\right)\right]\right]^\top}_{\text{equals to }0}\\
& \left[
\mathbb{E}_{\mathcal{M} \mid \mathcal{S}}\left[f^\sigma\left(\boldsymbol{x}; \boldsymbol{\hat{\theta}}_\mathcal{M}\right)\right] - f^*\left(\boldsymbol{x}\right)\right]\Bigg] \\
&+\mathbb{E}_{\boldsymbol{z}}\left[\left\|\mathbb{E}_{\mathcal{M} \mid \mathcal{S}}\left[f^\sigma\left(\boldsymbol{x}; \boldsymbol{\hat{\theta}}_\mathcal{M}\right)\right] - f^*\left(\boldsymbol{x}\right) \right\|_2^2 \right].\\
\end{aligned}
\end{equation*}

Applying the law of total expectation over the partition $\{\mathcal{Z}_i\}_{i=1}^{L_m}$, we decompose the global error into a weighted sum of local contributions:
\begin{equation*}
\begin{aligned}
&\mathcal{R}_{\mathcal{M} \mid \mathcal{S}}=\sum_{i=1}^{L_m} \mathbb{P}\left(\boldsymbol{z}\in \mathcal{Z}_i\right) \cdot \mathbb{E}_{\boldsymbol{z}\mid \mathcal{Z}_i}\left[\text{tr}\left[\text{Cov}_{\mathcal{M} \mid \mathcal{S}}\left[f^\sigma\left(\boldsymbol{x}; \boldsymbol{\hat{\theta}}_\mathcal{M}\right)\right]\right]\right]\\
&+\mathbb{E}_{\boldsymbol{z}}\left[\left\|\mathbb{E}_{\mathcal{M} \mid \mathcal{S}}\left[f^\sigma\left(\boldsymbol{x}; \boldsymbol{\hat{\theta}}_\mathcal{M}\right)\right] - f^*\left(\boldsymbol{x}\right)\right\|_2^2 \right].\\
\end{aligned}
\end{equation*}
\hfill$\square$
\section*{Relevant Lemmas and Proofs}
\newtheorem{lemma}{Lemma}[section]
\begin{lemma}
\label{lem:hausdorff}
Given a set $\mathcal{A} = \{a_1, a_2, \dots, a_l\}$ containing $l$ elements with $\text{diam}(\mathcal{A}) \leq m$, for any two sets $\mathcal{A}_1$ and $\mathcal{A}_2$ of sample sizes $S_1$ and $S_2$ respectively ($S_1\ge 0$ and $S_2\ge 0$), obtained by independent and identically distributed sampling from $\mathcal{A}$ with uniform distribution, and for any non-decreasing function $\varphi$, it holds that
\begin{equation*}
\mathbb{E}\left[\varphi\left(h\left(\mathcal{A}_1, \mathcal{A}_2\right)\right)\right]\le \left[\varphi\left(m\right)-\varphi\left(0\right)\right]\lambda\left(l,S_1,S_2\right)+ \varphi\left(0\right),
\end{equation*}
where 
\begin{equation*}
\begin{aligned}
\lambda\left(l,S_1,S_2\right)&=l\left[\left(1-l^{-1}\right)^{S_1}+\left(1-l^{-1}\right)^{S_2}\right.\\
& \left.-2\left(1-l^{-1}\right)^{S_2}\left(1-l^{-1}\right)^{S_1}\right].
\end{aligned}
\end{equation*}
\end{lemma}
\textit{Proof:}
It's easy to see that when $ l = 1 $, the Hausdorff distance between $\mathcal{A}_1$ and $\mathcal{A}_2$ must be 0, i.e., $\mathbb{P}\left(h\left(\mathcal{A}_1, \mathcal{A}_2\right) = 0\right) = 1$. In this case,
\begin{equation*}
\mathbb{E}\left[\varphi\left[h\left(\mathcal{A}_1, \mathcal{A}_2\right)\right]\right]=\varphi\left(0\right).
\end{equation*}
When $ l > 1 $, let event $E_i$ denote that ``$a_i \in \mathcal{A}_1$ and $ a_i \notin \mathcal{A}_2 $", and event $ F_i $ denote that ``$a_i \notin \mathcal{A}_1$ and $ a_i \in \mathcal{A}_2 $", then
\begin{equation*}
\begin{aligned}
\mathbb{P}\left(h\left(\mathcal{A}_1, \mathcal{A}_2\right) \ne 0\right)&=\mathbb{P}\left(\cup_{i=1}^l\left(E_i\cup F_i\right)\right)\\
&\le\sum_{i=1}^l\left[\mathbb{P}\left(E_i\right)+ \mathbb{P}\left(F_i\right)\right]\\
& =l\left[\left[1-\left(1-l^{-1}\right)^{S_1}\right]\left(1-l^{-1}\right)^{S_2}\right.\\
& \left.+\left[1-\left(1-l^{-1}\right)^{S_2}\right]\left(1-l^{-1}\right)^{S_1}\right]\\
& = \lambda\left(l,S_1,S_2\right),
\end{aligned}
\end{equation*}
using the fact that $\text{diam}(\mathcal A)\le m$, we have $h(\mathcal A_1,\mathcal A_2)\le m$, and therefore
\begin{equation*}
\begin{aligned}
&\mathbb{E}\left[h\left(\mathcal{A}_1, \mathcal{A}_2\right)\right]\\
&\le\mathbb{P}\left(h\left(\mathcal{A}_1, \mathcal{A}_2\right)\ne 0\right) \varphi\left(m\right)+\left[1-\mathbb{P}\left(h\left(\mathcal{A}_1, \mathcal{A}_2\right)\ne 0\right)\right]\varphi\left(0\right),\\
\end{aligned}
\end{equation*}
since the right-hand side of the inequality above is a monotonically increasing function of $\mathbb{P}\left(h\left(\mathcal{A}_1, \mathcal{A}_2\right)\ne 0\right)$, therefore
\begin{equation*}
\mathbb{E}\left[\varphi\left(h\left(\mathcal{A}_1, \mathcal{A}_2\right)\right)\right]\le \left[\varphi\left(m\right)-\varphi\left(0\right)\right]\lambda\left(l,S_1,S_2\right)+ \varphi\left(0\right).
\end{equation*}
\hfill$\square$

\begin{lemma}
\label{lemma:assump}
Under Assumption \ref{assump_1}, given any local partition $\{\mathcal{Z}_i\}_{i=1}^{L_m}$ of the sample space $\mathcal{Z}$ with each local unit has diameter $m$, the following inequality holds
\begin{equation}
\label{eq:pre-inequality}
\begin{aligned}
&\sum_{i=1}^{L_m}  \mathbb{P}\left(\boldsymbol{z}\in \mathcal{Z}_i\right) \cdot \mathbb{E}_{\boldsymbol{z}\mid \mathcal{Z}_i}\text{tr}\left[\text{Cov}_{\mathcal{M} \mid \mathcal{S}}\left[f^\sigma\left(\boldsymbol{x}; \boldsymbol{\hat{\theta}}_\mathcal{M}\right)\right]\right]\\
&\le\sum_{i\in\Omega_{-\emptyset}}  \mathbb{P}\left(\boldsymbol{z}\in \mathcal{Z}_i\right) \cdot \\
&\bigg[ \mathbb{E}_{\boldsymbol{z}\mid \mathcal{Z}_i}\mathbb{E}_{\mathcal{M} \mid \mathcal{S}}\left[ \left\|f^\sigma\left(\boldsymbol{x}; \boldsymbol{\hat{\theta}}_{\mathcal{M}_i}\right)-\mathbb{E}_{\mathcal{M} \mid \mathcal{S}}\left[f^\sigma\left(\boldsymbol{x}; \boldsymbol{\hat{\theta}}_{\mathcal{M}_i}\right)\right]\right\|_2^2\right]\\
&+2\gamma\mathbb{E}_{\boldsymbol{z}\mid \mathcal{Z}_i}\mathbb{E}_{\mathcal{M} \mid \mathcal{S}}\left[ \left\|f^\sigma\left(\boldsymbol{x}; \boldsymbol{\hat{\theta}}_{\mathcal{M}_i}\right)-\mathbb{E}_{\mathcal{M} \mid \mathcal{S}}\left[f^\sigma\left(\boldsymbol{x}; \boldsymbol{\hat{\theta}}_{\mathcal{M}_i}\right)\right]\right\|_2\right]\bigg]\\
&+4\gamma^2,
\end{aligned}
\end{equation}
where $\Omega_{-\emptyset} = \left\{i \in \left\{1, 2, \dots, L_m\right\} \mid \mathcal{Z}_i \cap\mathcal{S}  \ne \emptyset\right\}$ is the index set consisting of indices of all local regions $\mathcal{Z}_i$ that satisfy $\mathcal{Z}_i \cap \mathcal{S} \ne \emptyset$. 
\end{lemma}
\textit{Proof:}
We begin by addressing the first term in Eq. \eqref{eq:decomposition}. Denote $\mathcal{M}\cap\mathcal{Z}_i$ as $\mathcal{M}_i$, then for every region $\mathcal{Z}_i$, we have
\begin{equation*}
\begin{aligned}
&\mathbb{E}_{\boldsymbol{z}\mid \mathcal{Z}_i}\text{tr}\left[\text{Cov}_{\mathcal{M} \mid \mathcal{S}}\left[f^\sigma\left(\boldsymbol{x}; \boldsymbol{\hat{\theta}}_\mathcal{M}\right)\right]\right]\\
&=\mathbb{E}_{\boldsymbol{z}\mid \mathcal{Z}_i}\mathbb{E}_{\mathcal{M} \mid \mathcal{S}}\left[ \left\| f^\sigma\left(\boldsymbol{x}; \boldsymbol{\hat{\theta}}_\mathcal{M}\right)-\mathbb{E}_{\mathcal{M} \mid \mathcal{S}}\left[f^\sigma\left(\boldsymbol{x}; \boldsymbol{\hat{\theta}}_\mathcal{M}\right)\right]\right\|_2^2\right]\\
&=\mathbb{E}_{\boldsymbol{z}\mid \mathcal{Z}_i}\mathbb{E}_{\mathcal{M} \mid \mathcal{S}}\bigg[ \left\| f^\sigma\left(\boldsymbol{x}; \boldsymbol{\hat{\theta}}_\mathcal{M}\right)-f^\sigma\left(\boldsymbol{x}; \boldsymbol{\hat{\theta}}_{\mathcal{M}_i}\right)\right\|_2^2\\
&+\left\|f^\sigma\left(\boldsymbol{x}; \boldsymbol{\hat{\theta}}_{\mathcal{M}_i}\right)-\mathbb{E}_{\mathcal{M} \mid \mathcal{S}}\left[f^\sigma\left(\boldsymbol{x}; \boldsymbol{\hat{\theta}}_{\mathcal{M}_i}\right)\right]\right\|_2^2\\
&+\left\|\mathbb{E}_{\mathcal{M} \mid \mathcal{S}}\left[f^\sigma\left(\boldsymbol{x}; \boldsymbol{\hat{\theta}}_{\mathcal{M}_i}\right)-f^\sigma\left(\boldsymbol{x}; \boldsymbol{\hat{\theta}}_\mathcal{M}\right)\right]\right\|_2^2\\
&+2\left[f^\sigma\left(\boldsymbol{x}; \boldsymbol{\hat{\theta}}_\mathcal{M}\right)-f^\sigma\left(\boldsymbol{x}; \boldsymbol{\hat{\theta}}_{\mathcal{M}_i}\right)\right]^\top\\
&\left[f^\sigma\left(\boldsymbol{x}; \boldsymbol{\hat{\theta}}_{\mathcal{M}_i}\right)-\mathbb{E}_{\mathcal{M} \mid \mathcal{S}}\left[f^\sigma\left(\boldsymbol{x}; \boldsymbol{\hat{\theta}}_{\mathcal{M}_i}\right)\right]\right]\\
&+2\left[f^\sigma\left(\boldsymbol{x}; \boldsymbol{\hat{\theta}}_\mathcal{M}\right)-f^\sigma\left(\boldsymbol{x}; \boldsymbol{\hat{\theta}}_{\mathcal{M}_i}\right)\right]^\top\\
&\mathbb{E}_{\mathcal{M} \mid \mathcal{S}}\left[f^\sigma\left(\boldsymbol{x}; \boldsymbol{\hat{\theta}}_{\mathcal{M}_i}\right)-f^\sigma\left(\boldsymbol{x}; \boldsymbol{\hat{\theta}}_\mathcal{M}\right)\right]\\
&+2\left[f^\sigma\left(\boldsymbol{x}; \boldsymbol{\hat{\theta}}_{\mathcal{M}_i}\right)-\mathbb{E}_{\mathcal{M} \mid \mathcal{S}}\left[f^\sigma\left(\boldsymbol{x}; \boldsymbol{\hat{\theta}}_{\mathcal{M}_i}\right)\right]\right]^\top\\
&\mathbb{E}_{\mathcal{M} \mid \mathcal{S}}\left[f^\sigma\left(\boldsymbol{x}; \boldsymbol{\hat{\theta}}_{\mathcal{M}_i}\right)-f^\sigma\left(\boldsymbol{x}; \boldsymbol{\hat{\theta}}_\mathcal{M}\right)\right]\bigg].\\
\end{aligned}
\end{equation*}
Note that
\begin{equation*}
\begin{aligned}
&\mathbb{E}_{\mathcal{M} \mid \mathcal{S}}\bigg[\left[f^\sigma\left(\boldsymbol{x}; \boldsymbol{\hat{\theta}}_{\mathcal{M}_i}\right)-\mathbb{E}_{\mathcal{M} \mid \mathcal{S}}\left[f^\sigma\left(\boldsymbol{x}; \boldsymbol{\hat{\theta}}_{\mathcal{M}_i}\right)\right]\right]^\top\\
&\mathbb{E}_{\mathcal{M} \mid \mathcal{S}}\left[f^\sigma\left(\boldsymbol{x}; \boldsymbol{\hat{\theta}}_{\mathcal{M}_i}\right)-f^\sigma\left(\boldsymbol{x}; \boldsymbol{\hat{\theta}}_\mathcal{M}\right)\right]\bigg]\\
&=\mathbb{E}_{\mathcal{M} \mid \mathcal{S}}\bigg[\left[f^\sigma\left(\boldsymbol{x}; \boldsymbol{\hat{\theta}}_{\mathcal{M}_i}\right)-\mathbb{E}_{\mathcal{M} \mid \mathcal{S}}\left[f^\sigma\left(\boldsymbol{x}; \boldsymbol{\hat{\theta}}_{\mathcal{M}_i}\right)\right]\right]\bigg]^\top\\
&\mathbb{E}_{\mathcal{M} \mid \mathcal{S}}\left[f^\sigma\left(\boldsymbol{x}; \boldsymbol{\hat{\theta}}_{\mathcal{M}_i}\right)-f^\sigma\left(\boldsymbol{x}; \boldsymbol{\hat{\theta}}_\mathcal{M}\right)\right]\\
&=0.
\end{aligned}
\end{equation*}

Furthermore, with Assumption \ref{assump_1}, we obtain:
\begin{equation*}
\begin{aligned}
&\left[f^\sigma\left(\boldsymbol{x}; \boldsymbol{\hat{\theta}}_\mathcal{M}\right)-f^\sigma\left(\boldsymbol{x}; \boldsymbol{\hat{\theta}}_{\mathcal{M}_i}\right)\right]^\top\\
&\left[f^\sigma\left(\boldsymbol{x}; \boldsymbol{\hat{\theta}}_{\mathcal{M}_i}\right)-\mathbb{E}_{\mathcal{M} \mid \mathcal{S}}\left[f^\sigma\left(\boldsymbol{x}; \boldsymbol{\hat{\theta}}_{\mathcal{M}_i}\right)\right]\right]\\
&\le \gamma\cdot\left\|f^\sigma\left(\boldsymbol{x}; \boldsymbol{\hat{\theta}}_{\mathcal{M}_i}\right)-\mathbb{E}_{\mathcal{M} \mid \mathcal{S}}\left[f^\sigma\left(\boldsymbol{x}; \boldsymbol{\hat{\theta}}_{\mathcal{M}_i}\right)\right]\right\|_2.
\end{aligned}
\end{equation*}

By Jensen’s inequality, it follows that

\begin{equation*}
\begin{aligned}
&\left\|\mathbb{E}_{\mathcal{M} \mid \mathcal{S}}\left[f^\sigma\left(\boldsymbol{x}; \boldsymbol{\hat{\theta}}_{\mathcal{M}_i}\right)-f^\sigma\left(\boldsymbol{x}; \boldsymbol{\hat{\theta}}_\mathcal{M}\right)\right]\right\|_2\\
&\le\mathbb{E}_{\mathcal{M} \mid \mathcal{S}}\left[\left\|f^\sigma\left(\boldsymbol{x}; \boldsymbol{\hat{\theta}}_{\mathcal{M}_i}\right)-f^\sigma\left(\boldsymbol{x}; \boldsymbol{\hat{\theta}}_\mathcal{M}\right)\right\|_2\right]\\
&\le\gamma
\end{aligned}
\end{equation*}
and
\begin{equation*}
\begin{aligned}
&\left\|\mathbb{E}_{\mathcal{M} \mid \mathcal{S}}\left[f\left(\boldsymbol{x}; \boldsymbol{\hat{\theta}}_{\mathcal{M}_i}\right)-f\left(\boldsymbol{x}; \boldsymbol{\hat{\theta}}_\mathcal{M}\right)\right]\right\|_2^2\\
&\le \mathbb{E}_{\mathcal{M} \mid \mathcal{S}}\left[\left\|f\left(\boldsymbol{x}; \boldsymbol{\hat{\theta}}_{\mathcal{M}_i}\right)-f\left(\boldsymbol{x}; \boldsymbol{\hat{\theta}}_\mathcal{M}\right)\right\|_2^2\right]\\
&\le\gamma^2.
\end{aligned}
\end{equation*}

Thus
\begin{equation}
\label{eq:cov_bound_pre}
\begin{aligned}
&\mathbb{E}_{\boldsymbol{z}\mid \mathcal{Z}_i}\text{tr}\left[\text{Cov}_{\mathcal{M} \mid \mathcal{S}}\left[f^\sigma\left(\boldsymbol{x}; \boldsymbol{\hat{\theta}}_\mathcal{M}\right)\right]\right]\\
&\le \mathbb{E}_{\boldsymbol{z}\mid \mathcal{Z}_i}\mathbb{E}_{\mathcal{M} \mid \mathcal{S}}\left[ \left\|f^\sigma\left(\boldsymbol{x}; \boldsymbol{\hat{\theta}}_{\mathcal{M}_i}\right)-\mathbb{E}_{\mathcal{M} \mid \mathcal{S}}\left[f^\sigma\left(\boldsymbol{x}; \boldsymbol{\hat{\theta}}_{\mathcal{M}_i}\right)\right]\right\|_2^2\right]\\
&+2\gamma\mathbb{E}_{\boldsymbol{z}\mid \mathcal{Z}_i}\mathbb{E}_{\mathcal{M} \mid \mathcal{S}}\left[ \left\|f^\sigma\left(\boldsymbol{x}; \boldsymbol{\hat{\theta}}_{\mathcal{M}_i}\right)-\mathbb{E}_{\mathcal{M} \mid \mathcal{S}}\left[f^\sigma\left(\boldsymbol{x}; \boldsymbol{\hat{\theta}}_{\mathcal{M}_i}\right)\right]\right\|_2\right]\\
&+4\gamma^2.
\end{aligned}
\end{equation}

Note that when $\mathcal{S}\cap\mathcal{Z}_i=\emptyset$, it must holds that $\mathcal{M}\cap\mathcal{Z}_i=\emptyset$. In this case,
\begin{equation}
\label{eq:middle}
\begin{aligned}
&f^\sigma\left(\boldsymbol{x}; \boldsymbol{\hat{\theta}}_{\mathcal{M}_i}\right)-\mathbb{E}_{\mathcal{M} \mid \mathcal{S}}\left[f^\sigma\left(\boldsymbol{x}; \boldsymbol{\hat{\theta}}_{\mathcal{M}_i}\right)\right]\\
&=f^\sigma\left(\boldsymbol{x}; \boldsymbol{\hat{\theta}}_\emptyset\right)-\mathbb{E}_{\mathcal{M} \mid \mathcal{S}}\left[f^\sigma\left(\boldsymbol{x}; \boldsymbol{\hat{\theta}}_{\emptyset}\right)\right]\\
&=f^\sigma\left(\boldsymbol{x}; \boldsymbol{\hat{\theta}}_\emptyset\right)-f^\sigma\left(\boldsymbol{x}; \boldsymbol{\hat{\theta}}_\emptyset\right)=0,
\end{aligned}
\end{equation}
therefore,
\begin{equation*}
\begin{aligned}
&\sum_{i=1}^{L_m}  \mathbb{P}\left(\boldsymbol{z}\in \mathcal{Z}_i\right) \cdot \mathbb{E}_{\boldsymbol{z}\mid \mathcal{Z}_i}\text{tr}\left[\text{Cov}_{\mathcal{M} \mid \mathcal{S}}\left[f^\sigma\left(\boldsymbol{x}; \boldsymbol{\hat{\theta}}_\mathcal{M}\right)\right]\right]\\
&\le\sum_{i\in\Omega_{-\emptyset}} \mathbb{P}\left(\boldsymbol{z}\in \mathcal{Z}_i\right) \cdot \\
&\bigg[ \mathbb{E}_{\boldsymbol{z}\mid \mathcal{Z}_i}\mathbb{E}_{\mathcal{M} \mid \mathcal{S}}\left[ \left\|f^\sigma\left(\boldsymbol{x}; \boldsymbol{\hat{\theta}}_{\mathcal{M}_i}\right)-\mathbb{E}_{\mathcal{M} \mid \mathcal{S}}\left[f^\sigma\left(\boldsymbol{x}; \boldsymbol{\hat{\theta}}_{\mathcal{M}_i}\right)\right]\right\|_2^2\right]\\
&+2\gamma\mathbb{E}_{\boldsymbol{z}\mid \mathcal{Z}_i}\mathbb{E}_{\mathcal{M} \mid \mathcal{S}}\left[ \left\|f^\sigma\left(\boldsymbol{x}; \boldsymbol{\hat{\theta}}_{\mathcal{M}_i}\right)-\mathbb{E}_{\mathcal{M} \mid \mathcal{S}}\left[f^\sigma\left(\boldsymbol{x}; \boldsymbol{\hat{\theta}}_{\mathcal{M}_i}\right)\right]\right\|_2\right]\bigg]\\
&+4\gamma^2.
\end{aligned}
\end{equation*}
\hfill$\square$

\section*{Proof of theorem \ref{theo.1}}
We begin by analyzing the second term on the right-hand side of Eq. \eqref{eq:pre-inequality}. Consider the case where exactly $b$ samples from $\mathcal M$ fall into $\mathcal Z_i$, since the samples are independent and identically distributed, any choice of these $b$ samples is symmetric. Without loss of generality, denote the set consist of these $b$ samples as $\mathcal{M}_i^b$, and let $\mathbb{E}_{\mathcal{M}\mid \mathcal{S},\mathcal{M}_i^b}$ denote the conditional expectation given $\mathcal{M}_i=\mathcal{M}^b_i$. Then
\begin{equation}
\label{eq:er_decomp}
\begin{aligned}
&\mathbb{E}_{\mathcal{M} \mid \mathcal{S},\mathcal{M}_i^b}\left[ \left\|f^\sigma\left(\boldsymbol{x}; \boldsymbol{\hat{\theta}}_{\mathcal{M}_i}\right)-\mathbb{E}_{\mathcal{M} \mid \mathcal{S}}\left[f^\sigma\left(\boldsymbol{x}; \boldsymbol{\hat{\theta}}_{\mathcal{M}_i}\right)\right]\right\|_2\right]\\
&=\mathbb{E}_{\mathcal{M}_i \mid \mathcal{S},\mathcal{M}_i^b}\left[ \left\|f^\sigma\left(\boldsymbol{x}; \boldsymbol{\hat{\theta}}_{\mathcal{M}_i}\right)-\mathbb{E}_{\mathcal{M} \mid \mathcal{S}}\left[f^\sigma\left(\boldsymbol{x}; \boldsymbol{\hat{\theta}}_{\mathcal{M}_i}\right)\right]\right\|_2\right]\\
&=\mathbb{E}_{\mathcal{M}_i \mid \mathcal{S},\mathcal{M}_i^b}\bigg[ \bigg\|f^\sigma\left(\boldsymbol{x}; \boldsymbol{\hat{\theta}}_{\mathcal{M}_i}\right)-\\
&\sum_{j=0}^N\mathrm{P}_{i}^j\cdot\mathbb{E}_{\mathcal{M}_i \mid \mathcal{S},\mathcal{M}_i^j}\left[f^\sigma\left(\boldsymbol{x}; \boldsymbol{\hat{\theta}}_{\mathcal{M}_i}\right)\right]\bigg\|_2\bigg],
\end{aligned}
\end{equation}
where $\mathrm{P}_{i}^j=\mathbb{P}\left(\left|\mathcal{M}_i\right|=j\mid \mathcal{S}\right)$. The first equality above holds because the expectation’s interior only involves $\mathcal{M}_i$, and the samples in $\mathcal{M}$ are drawn independently. Therefore, given $\mathcal{M}_i = \mathcal{M}^b_i$, the outer expectation acts only on the samples in $\mathcal{M}^b_i$ (i.e., $\mathcal{M}_i$). The second equality follows by combining this fact with the law of total expectation. Here 
\begin{equation}
\begin{aligned}
&\left\|f^\sigma\left(\boldsymbol{x}; \boldsymbol{\hat{\theta}}_{\mathcal{M}_i}\right)-\sum_{j=0}^N\mathrm{P}_{i}^j\cdot\mathbb{E}_{\mathcal{M}_i \mid \mathcal{S},\mathcal{M}_i^j}\left[f^\sigma\left(\boldsymbol{x}; \boldsymbol{\hat{\theta}}_{\mathcal{M}_i}\right)\right]\right\|_2\\
&=\left\|\sum_{j=0}^N\mathrm{P}_{i}^j\cdot\left[f^\sigma\left(\boldsymbol{x}; \boldsymbol{\hat{\theta}}_{\mathcal{M}_i}\right)-\mathbb{E}_{\mathcal{M}_i \mid \mathcal{S},\mathcal{M}_i^j}\left[f^\sigma\left(\boldsymbol{x}; \boldsymbol{\hat{\theta}}_{\mathcal{M}_i}\right)\right]\right]\right\|_2\\
&=\left\|\sum_{j=0}^N\mathrm{P}_{i}^j\cdot\mathbb{E}_{\widetilde{\mathcal{M}}_i \mid \mathcal{S},\mathcal{M}_i^j}\left[f^\sigma\left(\boldsymbol{x}; \boldsymbol{\hat{\theta}}_{\mathcal{M}_i}\right)-f^\sigma\left(\boldsymbol{x}; \boldsymbol{\hat{\theta}}_{\widetilde{\mathcal{M}}_i}\right)\right]\right\|_2\\
& \le \sum_{j=0}^N\mathrm{P}_{i}^j\cdot\mathbb{E}_{\widetilde{\mathcal{M}}_i \mid \mathcal{S},\mathcal{M}_i^j}\left[\left\|f^\sigma\left(\boldsymbol{x}; \boldsymbol{\hat{\theta}}_{\mathcal{M}_i}\right)-f^\sigma\left(\boldsymbol{x}; \boldsymbol{\hat{\theta}}_{\widetilde{\mathcal{M}}_i}\right)\right\|_2\right],
\end{aligned}
\end{equation}
% 代入Eq. \eqref{eq:var_decomp}，有
substituting into Eq. \eqref{eq:er_decomp} yields
\begin{equation}
\label{eq:er_decomp_final}
\begin{aligned}
&\mathbb{E}_{\mathcal{M} \mid \mathcal{S},\mathcal{M}_i^b}\left[ \left\|f^\sigma\left(\boldsymbol{x}; \boldsymbol{\hat{\theta}}_{\mathcal{M}_i}\right)-\mathbb{E}_{\mathcal{M} \mid \mathcal{S}}\left[f^\sigma\left(\boldsymbol{x}; \boldsymbol{\hat{\theta}}_{\mathcal{M}_i}\right)\right]\right\|_2\right]\le\sum_{j=0}^N\mathrm{P}_{i}^j\\
&\cdot\mathbb{E}_{\mathcal{M}_i \mid \mathcal{S},\mathcal{M}_i^b}\mathbb{E}_{\widetilde{\mathcal{M}}_i \mid \mathcal{S},\mathcal{M}_i^j}\left[\left\|f^\sigma\left(\boldsymbol{x}; \boldsymbol{\hat{\theta}}_{\mathcal{M}_i}\right)-f^\sigma\left(\boldsymbol{x}; \boldsymbol{\hat{\theta}}_{\widetilde{\mathcal{M}}_i}\right)\right\|_2\right],
\end{aligned}
\end{equation}
% 当$\mathcal{M}_i^b$和$\mathcal{M}_i^j$都为$0$时，
when both $b$ and $j$ are zero,
\begin{equation}
\label{eq:both_zero}
\begin{aligned}
&\mathbb{E}_{\mathcal{M}_i \mid \mathcal{S},\mathcal{M}_i^b}\mathbb{E}_{\widetilde{\mathcal{M}}_i \mid \mathcal{S},\mathcal{M}_i^j}\left[\left\|f^\sigma\left(\boldsymbol{x}; \boldsymbol{\hat{\theta}}_{\mathcal{M}_i}\right)-f^\sigma\left(\boldsymbol{x}; \boldsymbol{\hat{\theta}}_{\widetilde{\mathcal{M}}_i}\right)\right\|_2\right]\\
&=\left\|f^\sigma\left(\boldsymbol{x}; \boldsymbol{\hat{\theta}}_{\emptyset}\right)-f^\sigma\left(\boldsymbol{x}; \boldsymbol{\hat{\theta}}_{\emptyset}\right)\right\|_2=0,
\end{aligned}
\end{equation}
if either $b = 0, j\ne 0$ or $b\ne 0,j = 0$, with Assumption \ref{assump_2}, and by the conception of Hausdorff limits \cite{kuratowski2014topology}, we have $h(\mathcal{M}_i^b, \mathcal{M}_i^j) = \infty$ and thus $\phi\left(h(\mathcal{M}_i^b, \mathcal{M}_i^j)\right) \le u$, therefore
\begin{equation}
\label{eq:only_one_zero}
\begin{aligned}
&\mathbb{E}_{\mathcal{M}_i \mid \mathcal{S},\mathcal{M}_i^b}\mathbb{E}_{\widetilde{\mathcal{M}}_i \mid \mathcal{S},\mathcal{M}_i^j}\left[\left\|f^\sigma\left(\boldsymbol{x}; \boldsymbol{\hat{\theta}}_{\mathcal{M}_i}\right)-f^\sigma\left(\boldsymbol{x}; \boldsymbol{\hat{\theta}}_{\widetilde{\mathcal{M}}_i}\right)\right\|_2\right]\\
&\le \max_{\mathcal{M}_i^b,\mathcal{M}_i^j}\left\|f^\sigma\left(\boldsymbol{x};\boldsymbol{\hat{\theta}}_{\mathcal{M}_i^b}\right)-f^\sigma\left(\boldsymbol{x}; \boldsymbol{\hat{\theta}}_{\mathcal{M}_i^j}\right)\right\|_2\\
&\le \phi\left(h\left(\mathcal{M}_i^b,\mathcal{M}_i^j\right)\right)\le u,
\end{aligned}
\end{equation}
whenever neither $b$ nor $j$ is zero, using the condition in Assumption \ref{assump_2} and applying Lemma \ref{lem:hausdorff} yields
\begin{equation}
\label{eq:both_non_zero}
\begin{aligned}
&\mathbb{E}_{\mathcal{M}_i \mid \mathcal{S},\mathcal{M}_i^b}\mathbb{E}_{\widetilde{\mathcal{M}}_i \mid \mathcal{S},\mathcal{M}_i^j}\left[\left\|f^\sigma\left(\boldsymbol{x}; \boldsymbol{\hat{\theta}}_{\mathcal{M}_i}\right)-f^\sigma\left(\boldsymbol{x}; \boldsymbol{\hat{\theta}}_{\widetilde{\mathcal{M}}_i}\right)\right\|_2\right]\\
&\le\mathbb{E}_{\mathcal{M}_i \mid \mathcal{S},\mathcal{M}_i^b}\mathbb{E}_{\widetilde{\mathcal{M}}_i \mid \mathcal{S},\mathcal{M}_i^j}\left[\phi\left(h\left(\mathcal{M}_i^b,\mathcal{M}_i^j\right)\right)\right]\\
&\le\left[\phi\left(m\right)-u_0\right]\lambda\left(l_i,b,j\right)+ u_0,
\end{aligned}
\end{equation}
where $l_i=\left|\mathcal{Z}_i\cap\mathcal{S}\right|$ denotes the number of samples from $\mathcal{S}$ that fall into the $i$-th local region, and 
\begin{equation*}
\begin{aligned}
\lambda\left(l_i,b,j\right)&=l_i\left[\left(1-l_i^{-1}\right)^{b}+\left(1-l_i^{-1}\right)^{j}\right.\\
& \left.-2\left(1-l_i^{-1}\right)^{j}\left(1-l_i^{-1}\right)^{b}\right].
\end{aligned}
\end{equation*}

Combining Eq. \eqref{eq:er_decomp_final}, \eqref{eq:both_zero}, \eqref{eq:only_one_zero}, and \eqref{eq:both_non_zero}, we obtain
\begin{equation}
\label{eq:cond_exp}
\begin{aligned}
&\mathbb{E}_{\mathcal{M} \mid \mathcal{S}}\left[ \left\|f^\sigma\left(\boldsymbol{x}; \boldsymbol{\hat{\theta}}_{\mathcal{M}_i}\right)-\mathbb{E}_{\mathcal{M} \mid \mathcal{S}}\left[f^\sigma\left(\boldsymbol{x}; \boldsymbol{\hat{\theta}}_{\mathcal{M}_i}\right)\right]\right\|_2\right]\\
& =\sum_{b=1}^N \mathrm{P}_i^b\mathbb{E}_{\mathcal{M} \mid \mathcal{S},\mathcal{M}_i^b}\left[ \left\|f^\sigma\left(\boldsymbol{x}; \boldsymbol{\hat{\theta}}_{\mathcal{M}_i}\right)-\mathbb{E}_{\mathcal{M} \mid \mathcal{S}}\left[f^\sigma\left(\boldsymbol{x}; \boldsymbol{\hat{\theta}}_{\mathcal{M}_i}\right)\right]\right\|_2\right]\\
& \le\sum_{b=0}^N \mathrm{P}_i^b\sum_{j=0}^N\mathrm{P}_{i}^j\\
& \cdot\mathbb{E}_{\mathcal{M}_i \mid \mathcal{S},\mathcal{M}_i^b}\mathbb{E}_{\widetilde{\mathcal{M}}_i \mid \mathcal{S},\mathcal{M}_i^j}\left[\left\|f^\sigma\left(\boldsymbol{x}; \boldsymbol{\hat{\theta}}_{\mathcal{M}_i}\right)-f^\sigma\left(\boldsymbol{x}; \boldsymbol{\hat{\theta}}_{\widetilde{\mathcal{M}}_i}\right)\right\|_2\right]\\
& \le 2\mathrm{P}_i^0\left(1-\mathrm{P}_i^0\right)u\\
&+\sum_{b=1}^N \mathrm{P}_i^b\sum_{j=1}^N\mathrm{P}_{i}^j\left[\left[\phi\left(m\right)-u_0\right]\lambda\left(l_i,b,j\right)+ u_0\right]\\
& = 2\mathrm{P}_i^0\left(1-\mathrm{P}_i^0\right)u+\left(1-\mathrm{P}_i^0\right)^2 u_0\\
& + \left[\phi\left(m\right)-u_0\right]\sum_{b=1}^N \mathrm{P}_i^b\sum_{j=1}^N\mathrm{P}_{i}^j\lambda\left(l_i,b,j\right).
\end{aligned}
\end{equation}

Let $\mathrm{p}_{i}=\frac{\sum_{j\in\mathcal{I}_i}\pi\left(\boldsymbol{z}_j\right)}{\sum_{k=1}^n \pi\left(\boldsymbol{z}_k\right)}$ be the probability that a sample $\boldsymbol{z}\in\mathcal{S}$ drawn according to the distribution $\pi\left(\boldsymbol{z}\right)$ belongs to the $i$-th local region $\mathcal{Z}_i$, where $\mathcal{I}_i=\left\{j\in\left\{1,2,\dots,n\right\}\mid \boldsymbol{z}_j\in\mathcal{Z}_i\cap\mathcal{S}\right\}$ is the index set of samples from $\mathcal{S}$ that belong to $\mathcal{Z}_i\cap\mathcal{S}$. Note that $\mathrm{P}_{i}^b=\binom{b}{N}\mathrm{p}_{i}^b\left(1-p_i\right)^{N-b}$, therefore
\begin{equation}
\label{eq:prob_sum}
\begin{aligned}
& \sum_{b=1}^N \mathrm{P}_i^b\sum_{j=1}^N\mathrm{P}_{i}^j\lambda\left(l_i,b,j\right)\\
& = l_i\left(1-\mathrm{P}_{i}^0\right)\sum_{b=1}^N\binom{b}{N}\left[\mathrm{p}_{i}\left(1-l_i^{-1}\right)\right]^b\left(1-\mathrm{p}_{i}\right)^{N-b}\\
& +l_i\left(1-\mathrm{P}_{i}^0\right)\sum_{j=1}^N\binom{j}{N}\left[\mathrm{p}_{i}\left(1-l_i^{-1}\right)\right]^j\left(1-\mathrm{p}_{i}\right)^{N-j}\\
& -2l_i \sum_{b=1}^N\binom{b}{N}\left[\mathrm{p}_{i}\left(1-l_i^{-1}\right)\right]^b\left(1-\mathrm{p}_{i}\right)^{N-b}\\
& \cdot \sum_{j=1}^N\binom{j}{N}\left[\mathrm{p}_{i}\left(1-l_i^{-1}\right)\right]^j\left(1-\mathrm{p}_{i}\right)^{N-j}.
\end{aligned}
\end{equation}

According to the binomial theorem, we have
\begin{equation}
\label{eq:binom}
\begin{aligned}
& \sum_{b=1}^N\binom{b}{N}\left[\mathrm{p}_{i}\left(1-l_i^{-1}\right)\right]^b\left(1-\mathrm{p}_{i}\right)^{N-b}\\
% & =\left[\mathrm{p}_{i}\left(1-l_i^{-1}\right)+1-\mathrm{p}_{i}\right]^N-\mathrm{P}_{i}^0\\
& =\left(1-\mathrm{p}_{i}l_i^{-1}\right)^N-\mathrm{P}_{i}^0.
\end{aligned}
\end{equation}

Substituting Eq. \eqref{eq:binom} into Eq. \eqref{eq:prob_sum}, we have
\begin{equation}
\label{eq:final_prob_sum}
\begin{aligned}
& \sum_{b=1}^N \mathrm{P}_i^b\sum_{j=1}^N\mathrm{P}_{i}^j\lambda\left(l_i,b,j\right)\\
& = 2l_i\left(1-\mathrm{P}_{i}^0\right)\left(1-\mathrm{p}_{i}l_i^{-1}\right)^N-2l_i\mathrm{P}_{i}^0\left(1-\mathrm{P}_{i}^0\right)\\
& -2l_i \left[\left(1-\mathrm{p}_{i}l_i^{-1}\right)^{N}-\mathrm{P}_{i}^0\right]^2\\
& = 2l_i\left[\left(1-\mathrm{p}_{i}l_i^{-1}\right)^N\left(1-\left(1-\mathrm{p}_{i}l_i^{-1}\right)^{N}\right)\right.\\
&\left.+\mathrm{P}_{i}^0\left(\left(1-\mathrm{p}_{i}l_i^{-1}\right)^{N}-1\right)\right].
\end{aligned}
\end{equation}

Substituting Eq. \eqref{eq:final_prob_sum} into Eq. \eqref{eq:cond_exp} and using the fact that $\mathrm{P}_{i}^0=\left(1-\mathrm{p}_{i}\right)^N$, we have
\begin{equation}
\label{eq:final_upper}
\begin{aligned}
&\mathbb{E}_{\mathcal{M} \mid \mathcal{S}}\left[ \left\|f^\sigma\left(\boldsymbol{x}; \boldsymbol{\hat{\theta}}_{\mathcal{M}_i}\right)-\mathbb{E}_{\mathcal{M} \mid \mathcal{S}}\left[f^\sigma\left(\boldsymbol{x}; \boldsymbol{\hat{\theta}}_{\mathcal{M}_i}\right)\right]\right\|_2\right]\\
& = \left(1-\left(1-\mathrm{p}_{i}\right)^N\right)\left[\left(1-\mathrm{p}_{i}\right)^N \left(2u-u_0\right) + \phi\left(0\right)\right] \\
& + 2l_i\left[\phi\left(m\right)-u_0\right]\left[\left(1-\mathrm{p}_{i}\right)^N\left(\left(1-\mathrm{p}_{i}l_i^{-1}\right)^{N}-1\right)\right.\\
& +\left.\left(1-\mathrm{p}_{i}l_i^{-1}\right)^N\left(1-\left(1-\mathrm{p}_{i}l_i^{-1}\right)^{N}\right)\right].
\end{aligned}
\end{equation}

Note that by taking $\varphi = \phi^2$ in Lemma \ref{lem:hausdorff}, a form similar to \eqref{eq:both_non_zero} can be derived:
\begin{equation}
\begin{aligned}
&\mathbb{E}_{\mathcal{M}_i \mid \mathcal{S},\mathcal{M}_i^b}\mathbb{E}_{\widetilde{\mathcal{M}}_i \mid \mathcal{S},\mathcal{M}_i^j}\left[\left\|f^\sigma\left(\boldsymbol{x}; \boldsymbol{\hat{\theta}}_{\mathcal{M}_i}\right)-f^\sigma\left(\boldsymbol{x}; \boldsymbol{\hat{\theta}}_{\widetilde{\mathcal{M}}_i}\right)\right\|_2\right]\\
&\le\mathbb{E}_{\mathcal{M}_i \mid \mathcal{S},\mathcal{M}_i^b}\mathbb{E}_{\widetilde{\mathcal{M}}_i \mid \mathcal{S},\mathcal{M}_i^j}\left[\phi\left(h\left(\mathcal{M}_i^b,\mathcal{M}_i^j\right)\right)\right]\\
&\le\left[\phi^2\left(m\right)-u_0\right]\lambda\left(l_i,b,j\right)+ u_0,
\end{aligned}
\end{equation}

Then, using a derivation similar to that from \eqref{eq:er_decomp} to \eqref{eq:final_upper}, we obtain
\begin{equation}
\label{eq:final_upper_2}
\begin{aligned}
&\mathbb{E}_{\mathcal{M} \mid \mathcal{S}}\left[ \left\|f^\sigma\left(\boldsymbol{x}; \boldsymbol{\hat{\theta}}_{\mathcal{M}_i}\right)-\mathbb{E}_{\mathcal{M} \mid \mathcal{S}}\left[f^\sigma\left(\boldsymbol{x}; \boldsymbol{\hat{\theta}}_{\mathcal{M}_i}\right)\right]\right\|_2^2\right]\\
& = \left(1-\left(1-\mathrm{p}_{i}\right)^N\right)\left[\left(1-\mathrm{p}_{i}\right)^N \left(2u^2-u_0^2\right) + u_0^2\right] \\
& + 2l_i\left[\phi^2\left(m\right)-u_0^2\right]\left[\left(1-\mathrm{p}_{i}\right)^N\left(\left(1-\mathrm{p}_{i}l_i^{-1}\right)^{N}-1\right)\right.\\
& +\left.\left(1-\mathrm{p}_{i}l_i^{-1}\right)^N\left(1-\left(1-\mathrm{p}_{i}l_i^{-1}\right)^{N}\right)\right].
\end{aligned}
\end{equation}

Note that here we only discuss the case where the local region $\mathcal{Z}_i$ satisfies $\mathcal{Z}_i \cap \mathcal{S} \ne \emptyset$. However, as shown in our analysis of Eqs. \eqref{eq:cov_bound_pre} and \eqref{eq:middle}, when $\mathcal{Z}_i \cap \mathcal{S} = \emptyset$, we have $\mathbb{E}_{\boldsymbol{z}\mid \mathcal{Z}_i}\text{tr}\left[\text{Cov}_{\mathcal{M} \mid \mathcal{S}}\left[f^\sigma\left(\boldsymbol{x}; \boldsymbol{\hat{\theta}}_\mathcal{M}\right)\right]\right] \le 4\gamma^2$. Combining this result and substituting Eqs. \eqref{eq:final_upper} and \eqref{eq:final_upper_2} into Eq. \eqref{eq:cov_bound_pre}, we obtain
\begin{equation*}
\begin{aligned}
&\mathbb{E}_{\boldsymbol{z}\mid \mathcal{Z}_i}\text{tr}\left[\text{Cov}_{\mathcal{M} \mid \mathcal{S}}\left[f^\sigma\left(\boldsymbol{x}; \boldsymbol{\hat{\theta}}_\mathcal{M}\right)\right]\right]\\
&\le \mathbbm{1}\left(\mathcal{Z}_i \cap \mathcal{S} \ne \emptyset\right) \bigg[ \left(1-\left(1-\mathrm{p}_{i}\right)^N\right) \Big[C_0+\left(1-\mathrm{p}_{i}\right)^N C_1\Big] \\
& + 2l_iC_2\left[\left(1-\mathrm{p}_{i}l_i^{-1}\right)^N-\left(1-\mathrm{p}_{i}\right)^N\right]\left(1-\left(1-\mathrm{p}_{i}l_i^{-1}\right)^{N}\right)\bigg]\\
&+4\gamma^2.
\end{aligned}
\end{equation*}
\hfill$\square$

\section*{Proof of corollary \ref{Corollary.1}}
Directly substituting Eqs. \eqref{eq:final_upper} and \eqref{eq:final_upper_2} into Eq. \eqref{eq:pre-inequality} gives:

\begin{equation*}
\begin{aligned}
&\sum_{i=1}^{L_m}  \mathbb{P}\left(\boldsymbol{z}\in \mathcal{Z}_i\right) \cdot \mathbb{E}_{\boldsymbol{z}\mid \mathcal{Z}_i}\text{tr}\left[\text{Cov}_{\mathcal{M} \mid \mathcal{S}}\left[f^\sigma\left(\boldsymbol{x}; \boldsymbol{\hat{\theta}}_\mathcal{M}\right)\right]\right]\\
&\le\sum_{i\in\Omega_{-\emptyset}}\mathbb{P}\left(\boldsymbol{z}\in \mathcal{Z}_i\right) \cdot \bigg[ \left(1-\left(1-\mathrm{p}_{i}\right)^N\right) \Big[C_0+\left(1-\mathrm{p}_{i}\right)^N C_1\Big] \\
& + 2l_iC_2\left[\left(1-\mathrm{p}_{i}l_i^{-1}\right)^N-\left(1-\mathrm{p}_{i}\right)^N\right]\left(1-\left(1-\mathrm{p}_{i}l_i^{-1}\right)^{N}\right)\bigg]\\&+4\gamma^2.
\end{aligned}
\end{equation*}
\hfill$\square$

\vfill

\end{document}